%% file: ms.tex
\documentclass[runningheads]{llncs}
\pdfoutput=1
\input{tex_files/preamble.tex}
\input{tex_files/commands.tex}

\begin{document}
\pagestyle{headings}
\mainmatter

\title{Unsupervised Part Discovery by \\ Unsupervised Disentanglement}

\titlerunning{Unsupervised Part Discovery}
\authorrunning{S. Braun et al.}
\author{
    Sandro Braun ~~~ Patrick Esser ~~~ Bj\"orn Ommer
}
\institute{
    Heidelberg Collaboratory for Image Processing / IWR, Heidelberg University \\
    \small \{first name\}.\{last name\}@iwr.uni-heidelberg.de
}

\maketitle

\input{tex_files/paper.tex}

\clearpage
\bibliographystyle{splncs04}
\bibliography{ms}
\clearpage
\appendix
\input{tex_files/appendix.tex}
\end{document}

%% file: tex_files/preamble.tex
\usepackage{makeidx}
\usepackage{graphicx}
\usepackage{bmpsize}
\usepackage{amsmath,amssymb} %
\usepackage{pifont} %
\usepackage{xcolor}

\usepackage{tikz}
\usepackage{tikzscale}
\usepackage{pgfplots}
\pgfplotsset{compat=1.14}
\pgfdeclarelayer{bg1}    %
\pgfdeclarelayer{bg2}
\pgfsetlayers{bg2,bg1,main}
\usetikzlibrary{bayesnet}

\usepackage[font=small]{caption}
\usepackage[font=small]{subcaption} %
\usepackage{silence}
\usepackage{siunitx}
\usepackage{booktabs} %
\usepackage{placeins} %
\usepackage{epsfig}
\usepackage{cite}
\usepackage{multirow}
\usepackage{siunitx}
\usepackage{multibib}
\usepackage{super_math}
\usepackage[bookmarks=false]{hyperref}

\definecolor{deepblue}{rgb}{0,0,0.5}
\definecolor{deepred}{rgb}{0.6,0,0}
\definecolor{deepgreen}{rgb}{0,0.5,0}

\graphicspath{
    {./figures/svg/}
}

%% file: tex_files/commands.tex
\newcommand{\tabref}[1]{Tab.~\ref{#1}}
\newcommand{\figref}[1]{Fig.~\ref{#1}}
\newcommand{\secref}[1]{Sec.~\ref{#1}}

\newcommand{\citep}{\cite} %

\def\pennaction#{Pennaction}
\def\deepfashion#{DeepFashion}
\def\exercise#{Exercise}
\def\cub#{CUB}
\def\celeba#{celeba}

\newcommand{\fnum}[1]{\num[round-mode=places,round-precision=3]{#1}}
\newcommand{\bfnum}[1]{\num[round-mode=places,round-precision=3,math-rm=\mathbf]{#1}}
\newcommand{\ufnum}[1]{\underline{\num[round-mode=places,round-precision=3]{#1}}}

\DeclareMathOperator{\trace}{tr}

\DeclareMathOperator*{\argmax}{arg\,max}

\newcommand{\entropy}[1]{H({#1})}
\newcommand{\loss}[1]{\mathcal{L}_{#1}}
\newcommand{\lossweight}[1]{\lambda_{#1}}

\newcommand{\cmark}{\ding{51}}
\newcommand{\xmark}{\ding{55}}
\def\setreal{\mathbb{R}}

\DeclareMathOperator{\kl}{KL}

\newcommand{\kldiv}[2]{\kl \left( {#1} \middle\| {#2} \right) }

\DeclareFontFamily{U}{skulls}{}
\DeclareFontShape{U}{skulls}{m}{n}{ <-> skull }{}
\newcommand{\skull}{\text{\usefont{U}{skulls}{m}{n}\symbol{'101}}}

\newcommand{\E}[2]{
\mathrm{E}_{#1} ~ {#2}
}
\newcommand{\encoderalpha}[1]{
    E_\alpha
}
\newcommand{\encoderpi}[1]{
    E_\pi
}
\newcommand{\maskdecoder}[1]{
    D_{S}
}

\newcommand{\centered}[1]{\begin{tabular}{c} #1 \end{tabular}}
\newcommand{\normal}[2]{\mathcal{N}\left({#1},{#2}\right)}
\newcommand{\alphapart}[2]{\alpha^{#1}_{#2}}
\newcommand{\dkl}{\mathrm{KL}}

%% file: tex_files/paper.tex
\begin{abstract}
    We address the problem of discovering part segmentations of articulated objects without supervision.
    In contrast to keypoints, part segmentations provide information about part localizations on the level of individual pixels.
    Capturing both locations and semantics, they are an attractive target for supervised learning approaches.
    However, large annotation costs limit the scalability of supervised algorithms to other object categories than humans.
    Unsupervised approaches potentially allow to use much more data at a lower cost.
    Most existing unsupervised approaches focus on learning abstract representations to be refined with supervision into the final representation.
    Our approach leverages a generative model consisting of two disentangled representations for an object's shape and appearance and a latent variable for the part segmentation.
    From a single image, the trained model infers a semantic part segmentation map.
    In experiments, we compare our approach to previous state-of-the-art approaches and observe significant gains in segmentation accuracy and shape consistency \footnote{
        Code available at \url{https://compvis.github.io/unsupervised-part-segmentation}
    }.
    Our work demonstrates the feasibility to discover semantic part segmentations without supervision.
\end{abstract}
\section{Introduction}

Instances of articulated objects such as humans, birds and dogs differ in their articulation (different pose) and also show
different colors and textures (appearance).
Despite those large variations in articulation and appearance, humans are able to establish correspondences between individual parts across instances.

\begin{figure}[t]
    \null\hfill
    \begin{subfigure}[b]{0.35\textwidth}
        \centering
        \def\ww{0.4}
        \footnotesize
        \begin{tabular}{ccc}
            ~
             & Instance 1
             & Instance 2
            \\
            \parbox[c]{5mm}{{ \rotatebox[origin=b]{90}{ Image }}}
             & \parbox[c]{\ww\linewidth}{\includegraphics[width=1.0\linewidth]{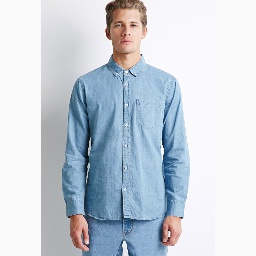}}
             & \parbox[c]{\ww\linewidth}{\includegraphics[width=1.0\linewidth]{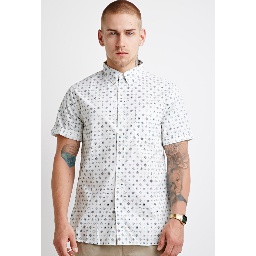}}
            \\
            \parbox[c]{5mm}{{ \rotatebox[origin=b]{90}{ Segmentation }}}
             & \parbox[c]{\ww\linewidth}{\includegraphics[width=1.0\linewidth]{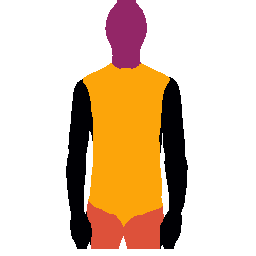}}
             & \parbox[c]{\ww\linewidth}{\includegraphics[width=1.0\linewidth]{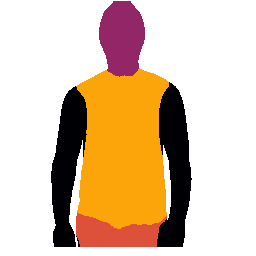}}
        \end{tabular}
        \caption{
            Two people with similar poses, $\pi_1 \simeq \pi_2$, but different
            appearances $\alpha_1 \neq \alpha_2$.
            Semantic segmentations $S_1, S_2$ are unaffected
            by appearance variations, i.e. $S_1 \simeq S_2$, and thus independent
            thereof.
        }
        \label{fig:methodappearanceinvariancsa}
    \end{subfigure}
    \hfill
    \begin{subfigure}[b]{0.55\textwidth}
        \centering
        \footnotesize
        \includegraphics[height=3.5cm]{figures/tikz/methodappearanceinvariances.tikz}
        \caption{
            In our model, the joint distribution over images $x$, segmentations
            $S$, poses $\pi$, and appearances $\alpha$ factorizes into $p(x, S,
                \pi , \alpha ) = p(x | S, \alpha) p(S | \pi) p(\pi) p(\alpha)$.
            Thus, $S$ is independent of $\alpha$ and dependent on $\pi$.
            While $\pi$ is a latent representation of pose, $S$ is a semantic segmentation.
        }
        \label{fig:methodappearanceinvariancsb}
    \end{subfigure}
    \hfill\null
    \caption{\textbf{A probabilistic model for unsupervised part discovery.} As
        illustrated in a), semantic segmentations are appearance independent, which
        is reflected in the structure of our probabilistic model shown in b).
    }
    \label{fig:methodappearanceinvariancsb}
\end{figure}

For example, consider two persons wearing different outfits as in
\figref{fig:methodappearanceinvariancsa}.
One is wearing a plain, blue shirt,
the other one is wearing a dotted, white T-shirt.
In the first case, arms and chest share the same appearance, thus information about appearances cannot be used to identify the parts.
In the second case, arms and chest have different appearances, thus information about appearances could be used to identify the parts.

Most previous approaches for learning part segmentations are based on supervised learning.
While this can lead to good performance on a narrow set of object classes,
especially that of humans \cite{guler2018densepose},
it requires to build a large dataset for each object of interest.
To overcome this limitation, we require methods that discover parts and their
segmentations solely from observing the data, i.e. we need unsupervised
approaches.

Previous works on unsupervised keypoint discovery \cite{zhang2018unsupervised,
    jakab2018unsupervised, lorenz2019} produce semantic keypoints which could
provide information about parts. However, as we show in our experiments, even
when combined with image intensity information to estimate the shape of parts,
inferring pixel-wise localizations of parts from keypoints remains ambiguous.
An essential ingredient of keypoint-based approaches is the built-in
low-dimensional bottleneck which encourages compression and hence learning.
The keypoints are represented through heatmaps of spatially normalized activations,
which encourages well localized activations i.e. keypoints.
In contrast, a segmentation of parts has roughly the same dimensionality as the
image itself and allows arbitrary shapes of the segmented parts.
Thus we cannot use the segmentation as a built-in
bottleneck and must find a different way to enforce the bottleneck.

To learn parts and their segmentation unsupervised,
we propose a probabilistic generative model with
three hidden variables.
We use two low-dimensional, continuous variables, which are independent of each
other, to disentangle the instance-specific appearance, from
the instance-invariant shape.
The third variable is a
high-dimensional discrete variable to model the support of parts, hence a segmentation.
It is a descendant of the appearance-independent shape variable to ensure independence of instance specific appearance.
We show how the mask can be efficiently learned in a
variational inference framework assuming suitable priors.
Overall,
our approach learns to infer a semantic part segmentation map from a
single image by learning from a stream of video frames or
from pairs of synthetically transformed images.

In experiments on multiple datasets of humans and birds, our method
is able to discover parts within the image that are consistent across
instances.
We compare intersection-over-union metrics (IOU) of our approach to those
obtained from previous methods on keypoint learning and observe improvements in
two out of three datasets of humans and on the dataset of birds.
In addition, the generative nature of our approach enables part-based
appearance transfers where it outperforms both pose supervised and
keypoint-based unsupervised approaches in terms of shape consistency.

\section{Background}

\paragraph{Disentangled Representation Learning}

To learn more meaningful representations,
\cite{rubenstein2018learning,
    eastwood2018a} build upon the Variational Autoencoder (VAE) \citep{VAE,
    VAE2} to encourage disentanglement. However, non-identifiability issues
\citep{Hyvrinen1999NonlinearIC, locatello2019challenging} suggest that
additional information is required to obtain well-defined factors.

\cite{kingma2014semi} demonstrated a factorization into style and content of
digits using a conditional variant of the VAE.
Motivated by Generative
Adversarial Networks (GANs) \citep{gan}, \cite{mathieu2016disentangling}
adds a discriminator to this architecture.
Using videos,
\cite{denton2017unsupervised} uses a classification problem to obtain
disentangled representations of the temporally varying factors and
its stationary factors.
This approach is closely related to estimating
and minimizing the Mutual Information of two factors
\citep{belghazi2018mine} by defining the joint distribution of the two
factors through samples from the same video.

\paragraph{Localized Representation Learning}

Image segmentation is a well studied problem in computer vision.
The seminal
work of \cite{Mumford1989OptimalApproximationsPiecewise} introduced a variational formulation to
approximate images by piecewise constant functions with regularized edge
length.
Superpixel approaches \citep{ren2003learning} group nearby pixels
according to their similarity and obtain oversegmentations of an
image.
\cite{arbelaez2006boundary} combines a hierarchy of segmentations
with contour detection to improve results.
However, these
methods rely on low-level image features and cannot account for semantic
similarity.

Co-segmentation assumes the
availability of a large number of examples showing the object to be
segmented.
The ability of this paradigm to learn from such a weak source of
information resulted in many different approaches \citep{lu2019} ranging
from graphical models \citep{vicente2011object} to deep generative models
\citep{singh2019finegan}.
But their underlying assumption that the object to
be segmented is salient limits them to masks of a single object, whereas our
method learns multiple semantic parts, with part-wise correspondences across
instances.

\paragraph{Unsupervised Part Discovery}

Part based models have been extensively studied
\cite{monroyBoundingBoxesLearningObject2012,
    uferDeepSemanticFeature2017a,
    yarlagaddaMeaningfulContoursDiscriminative2012,
    rubioGenerativeRegularizationLatent2015
}. Recent works demonstrated the ability to discover semantic keypoints without
supervision. Based on the differentiable score-map to keypoint layer of
\cite{yi2016lift}, \cite{Thewlis2017UnsupervisedLearningObject} learns keypoints which are
stable under synthetic image transformations by enforcing an equivariance
principle.
\cite{zhang2018unsupervised} integrates this principle into an
autoencoder framework.
\cite{jakab2018unsupervised} uses a reconstruction
task with two images from the same video, instead of synthetic
transformations.
\cite{lorenz2019} makes the representation more expressive
by considering ellipses instead of circles for keypoints.
However, in all cases the intermediate representation of keypoints is crucial,
We obtain pixel-accurate
part memberships, whereas above approaches can only give a rough heatmap of
part localizations.
In addition, our approach can handle occlusions robustly which we demonstrate in our experiments.

\section{Approach}

\begin{figure}[tb]
    \centering
    \footnotesize
    \def\svgwidth{\textwidth}
    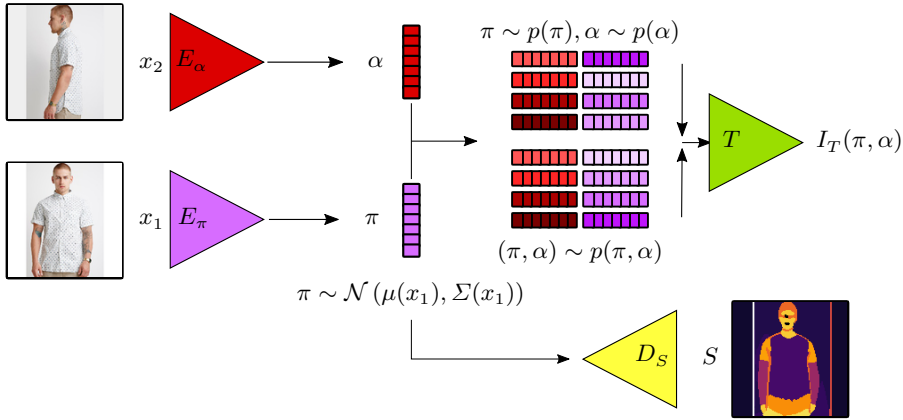
    \caption{
        \textbf{Learning Appearance Independent Segmentations}.
        To learn segmentations $S$ independent to appearance variation $\alpha$, we first disentangle
        a global representation for shape $\pi$ and appearance $\alpha$. The disentanglement is achieved
        through a variational and an adversarial constraint.
    }
    \label{fig:methodsegmentations}
\end{figure}

We have an image $x$ depicting an object $o$ composed of $N$ object parts $o_1, \dots, o_{N}$.
We would like to build a model that learns about those object parts
and assigns each location in the image to its corresponding object part, thus a part segmentation.
Without supervision for part segmentations, we rely on a generative approach by
looking for the segmentation $S^{\ast}$ that explains the image $x$ best.
Using Bayes rule, we can rewrite this as follows:
\begin{align}
    S^{\ast} = \argmax_{S} p(S | x) = \argmax p(x | S) p(S).
    \label{eq:mapmotivation}
\end{align}
The likelihood $p(x | S)$ measures if the segmentation can describe the image well enough
and the prior $p(S)$ measures if $S$ is a suitable candidate for a segmentation.
We now motivate suitable choices for the priors of $S$ for part learning.

\subsection{Appearance Independence of Segmentations}

\begin{figure}[tb]
    \null\hfill
    \begin{subfigure}[t]{0.47\textwidth}
        \centering
        \footnotesize
        \includegraphics[height=4cm, axisratio=1]{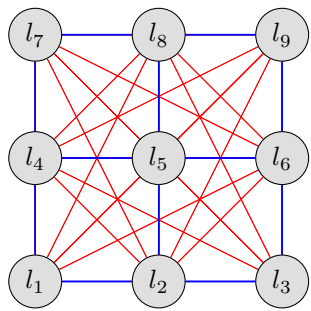}
        \caption{
            \textbf{Gaussian Markov Random Field}.
            Without any prior assumptions, any image pixel is dependent on any other pixel, (\textcolor{red}{\textbf{dense}} connectivity).
            In a GMRF, we only allow adjacent pixels to interact (\textcolor{blue}{\textbf{sparse}} connectivity).
        }
        \label{fig:metodsegmentationpriorsa}
    \end{subfigure}
    \hfill
    \begin{subfigure}[t]{0.47\textwidth}
        \centering
        \footnotesize
        \includegraphics[height=4cm, width=\textwidth]{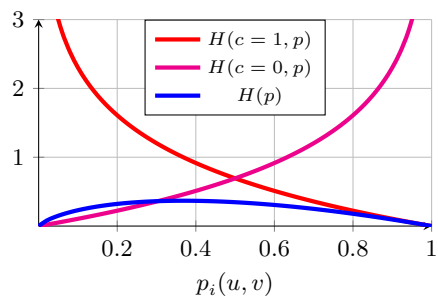}
        \caption{\textbf{Entropy Regularization}.
            To keep the learned segmentation $S$ close to a categorical distribution, we regularize the entropy of part probabilities $p_i(u, v)$.
        }
        \label{fig:metodsegmentationpriorsb}
    \end{subfigure}
    \caption{\textbf{Segmentation Priors}. We assume two priors for our segmentation model.}
    \label{fig:metodsegmentationpriors}
    \hfill\null
\end{figure}

Take two people spontaneously striking the same pose as depicted in \figref{fig:methodappearanceinvariancsa}.
The two people have the same pose $\pi$, but different appearances $\alpha_1$ and $\alpha_2$.
This generates two images $x_1$ and $x_2$ and we infer corresponding part segmentations $S_1$ and $S_2$.
Intuitively, the part segmentations $S_i$ are independent of the variation of individual appearances, and as those people share the same poses, clearly the part segmentation of the
image will be the same, i.e. $S_1=S_2$.
We argue that we can exploit this independence by modifying the image generation process so that segmentations are a result of pose and shape.
We now consider $\alpha$ and $\pi$ as random variables. In a graphical model sense, the joint distribution over poses, appearances and segmentations should factorize as depicted in \figref{fig:methodappearanceinvariancsb}:
$p(x, S, \pi , \alpha ) = p(x | S, \alpha) p(S | \pi) p(\pi)$.
Note that this directly reveals the corresponding motivation in \eqref{eq:mapmotivation}.
If we had access to the underlying shape and appearance variables that generate images $x$, we know that the part segmentation $S$ must be dependent on shape $\pi$.
In practice, $\pi$ and $\alpha$ are hidden variables and we must learn to infer
them from observations $x$.

\subsection{Learning Appearance Independent Segmentations}
\label{sec:learning_appearance_independent_segmentations}

We now explain how we achieve a disentangled representation of shape and appearance.
Let $x_i \sim (\alpha_i, \pi_i)$ express that $\alpha_i$ and $\pi_i$ were the factors generating the image $x_i$.
We then sample $x_1 = (\alpha, \pi_1)$ and $x_2 = (\alpha, \pi_2)$ from the dataset. In practice, this means that we need a pair of images depicting the same object but with varied poses.
To infer the latent variables, we use two encoders.
\begin{align}
    E_\alpha ~:~ & \setreal^{\dim(x)} \to \setreal^{\dim(\alpha)},  x \mapsto \alpha,
    ~~~
    E_\alpha(x_2) = \alpha                                                            \\
    E_\pi ~:~    & \setreal^{\dim(x)} \to \setreal^{\dim(\pi)}, x \mapsto \pi,
    ~~~
    E_\pi(x_1) = \pi
\end{align}
Here, $\alpha$ and $\pi$ are simple low-dimensional latent variables, each represented by a vector.
Please refer to the appendix for implementation details.
To keep $\pi$ independent of $\alpha$, we simply keep $\pi$ close to a standard
normal distribution in a variational framework, i.e.
$p(\pi) = \normal{0}{I}$ and $q(\pi | x_1) = \normal{\mu(x_1)}{\Sigma(x_1)}$.
However, this is not sufficient to guarantee that $p(\pi)$ is factorized into semantically consistent parts.
To give an example, the model could also learn to factorize parts based on their average color, i.e. all blue parts and all red parts.
To prevent this, we add an additional adversarial constraint that limits the mutual information  between shape and appearance $I(\alpha, \pi)$.
Following recent works on mutual information estimation
\cite{Belghazi2018MINEMutualInformation,Poole2019VariationalBoundsMutual,Mescheder2017AdversarialVariationalBayes},
we achieve this through an adversary $T$, which is a simple classifier trained
with the following objective
\begin{align}
    \max_{T}~
     & \E{
        (\pi, \alpha) \sim p(\pi, \alpha)
    }{
        \log \left(
        \sigma (
        T (
        \pi, \alpha
        )
        )
        \right)
    } +
    \\
     & \E{
        \pi \sim p(\pi), \alpha \sim p(\alpha)
    }{
        \log \left(
        1 - \sigma (
        T (
        \pi, \alpha
        )
        )
        \right)
    }
\end{align}
Here, $\sigma (x)$ denotes the sigmoid activation.
Intuitively, this means that we sample a batch of
$B$ image pairs,
$\{x_1^i, x_2^i\}$
$i = 1, \dots, B$, from the dataset and map them through the encoders,
$\alpha = \encoderalpha{}(x_2)$ and
$\pi = \encoderpi{}(x_1)$.
This gives us a batch of samples from the joint distribution
$(\pi, \alpha)_i \sim p(\pi, \alpha)$, $i = 1, \dots, B$.
We then randomly permute the order of $\{ \alpha_i \}$ within the batch to obtain a batch of samples from the marginal distribution $\pi \sim p(\pi), \alpha \sim p(\alpha)$.

\begin{figure}[tb]
    \centering
    \footnotesize
    \def\svgwidth{\textwidth}
    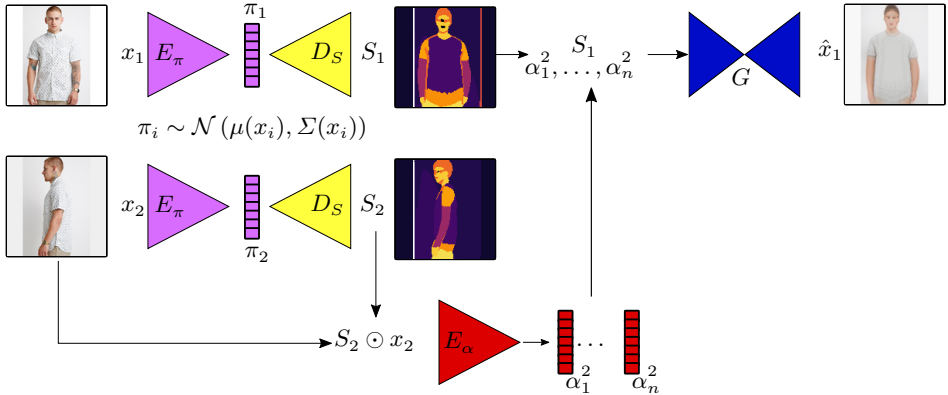
    \caption{
        \textbf{Complete Method}.
        First, we sample images $x_1$, $x_2$ from the dataset and infer their segmentations $S_1$ and $S_2$.
        We extract part based descriptors for appearance $\alphapart{2}{1}, \dots ,\alphapart{2}{N}$ from $x_2$ by  masking out each part using $S_2$ and mapping it into appearance space using $\encoderalpha{}$.
        We then build a likelihood model for $x_1$ based on $S_1$ and $\alphapart{2}{1}, \dots, \alphapart{2}{N}$.
    }
    \label{fig:methodcomplete}
\end{figure}

The procedure is depicted in \figref{fig:methodsegmentations}.
Note that the procedure is not a classical image discriminator as used in a GAN \cite{gan} training,
but rather a neural mutual information estimator \cite{belghazi2018mine,Esser2019UnsupervisedRobustDisentangling}.
One can derive that in the limit, the adversary converges to an estimate of the mutual information.
We thus term
$
    \E{
        (\pi, \alpha) \sim p(\pi, \alpha)
    }{
        T (\pi, \alpha)
    } = I_{T}(\pi, \alpha) = \widehat{I}(\pi, \alpha) $ an estimate for the mutual information of our disentangled representation.
This summarizes the objectives used to train the encoders.
\begin{align}
    \encoderpi{} ~:~
     & \min~ \loss{rec}
    + \lossweight{\text{variational}} \kldiv{
        q(\pi | x)
    }{
        p(\pi)
    }
    + \lossweight{\text{adversarial}} I_T(\pi, \alpha)
    \\
    \encoderalpha{} ~:~
     & \min~ \loss{rec}
\end{align}
Here $\loss{rec}$ is a reconstruction likelihood, such as a $\mathcal{L}_2$ loss or a perceptual loss between the original and the reconstructed image.
$\loss{rec}$ will be explained in more detail in \secref{sec:part_based_image_generation}.
In practice, we rely on the adaptive regularization scheme proposed in \cite{Esser2019UnsupervisedRobustDisentangling}.

Having a disentangled representation for shape and appearance, we can finally infer segmentations $S$ given shapes $\pi$ using a simple decoder model $\maskdecoder{}$.
The full procedure of disentanglement and inference for segmentations is depicted in \figref{fig:methodsegmentations}.
However, without further prior knowledge, it is in general not clear that $\maskdecoder{}$ will produce what resembles part segmentation under a common prior.
We therefore need to formulate suitable priors for $S$ to achieve the desired result.

\begin{figure}[t]
    \def\ww{0.082}
    \centering
    \begin{tabular}{l|ccc|ccc|ccc}
        ~
         & \multicolumn{3}{c}{\deepfashion{}}
         & \multicolumn{3}{c}{\exercise{}}
         & \multicolumn{3}{c}{\pennaction{}}
        \\
        \hline
        Input
         & \parbox[c]{\ww\linewidth}{\includegraphics[width=1.0\linewidth]{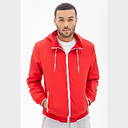}}
         & \parbox[c]{\ww\linewidth}{\includegraphics[width=1.0\linewidth]{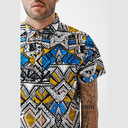}}
         & \parbox[c]{\ww\linewidth}{\includegraphics[width=1.0\linewidth]{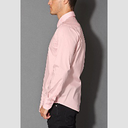}}
         & \parbox[c]{\ww\linewidth}{\includegraphics[width=1.0\linewidth]{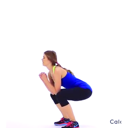}}
         & \parbox[c]{\ww\linewidth}{\includegraphics[width=1.0\linewidth]{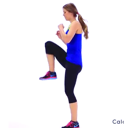}}
         & \parbox[c]{\ww\linewidth}{\includegraphics[width=1.0\linewidth]{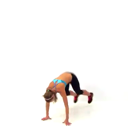}}
         & \parbox[c]{\ww\linewidth}{\includegraphics[width=1.0\linewidth]{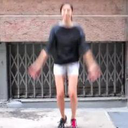}}
         & \parbox[c]{\ww\linewidth}{\includegraphics[width=1.0\linewidth]{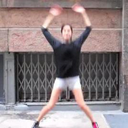}}
         & \parbox[c]{\ww\linewidth}{\includegraphics[width=1.0\linewidth]{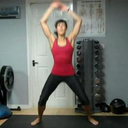}}
        \\
        \midrule
        \cite{zhang2018unsupervised}
         & \parbox[c]{\ww\linewidth}{\includegraphics[width=1.0\linewidth]{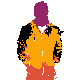}}
         & \parbox[c]{\ww\linewidth}{\includegraphics[width=1.0\linewidth]{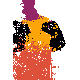}}
         & \parbox[c]{\ww\linewidth}{\includegraphics[width=1.0\linewidth]{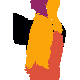}}
         & \parbox[c]{\ww\linewidth}{\includegraphics[width=1.0\linewidth]{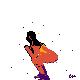}}
         & \parbox[c]{\ww\linewidth}{\includegraphics[width=1.0\linewidth]{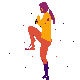}}
         & \parbox[c]{\ww\linewidth}{\includegraphics[width=1.0\linewidth]{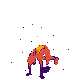}}
         & \parbox[c]{\ww\linewidth}{\includegraphics[width=1.0\linewidth]{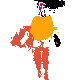}}
         & \parbox[c]{\ww\linewidth}{\includegraphics[width=1.0\linewidth]{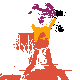}}
         & \parbox[c]{\ww\linewidth}{\includegraphics[width=1.0\linewidth]{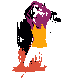}}
        \\
        \midrule
        \cite{jakab2018unsupervised}
         & \parbox[c]{\ww\linewidth}{\includegraphics[width=1.0\linewidth]{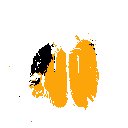}}
         & \parbox[c]{\ww\linewidth}{\includegraphics[width=1.0\linewidth]{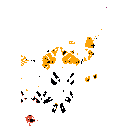}}
         & \parbox[c]{\ww\linewidth}{\includegraphics[width=1.0\linewidth]{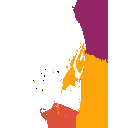}}
         & \parbox[c]{\ww\linewidth}{\includegraphics[width=1.0\linewidth]{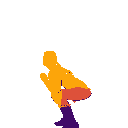}}
         & \parbox[c]{\ww\linewidth}{\includegraphics[width=1.0\linewidth]{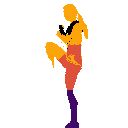}}
         & \parbox[c]{\ww\linewidth}{\includegraphics[width=1.0\linewidth]{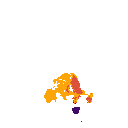}}
         & \parbox[c]{\ww\linewidth}{\includegraphics[width=1.0\linewidth]{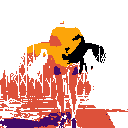}}
         & \parbox[c]{\ww\linewidth}{\includegraphics[width=1.0\linewidth]{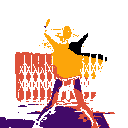}}
         & \parbox[c]{\ww\linewidth}{\includegraphics[width=1.0\linewidth]{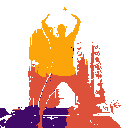}}
        \\
        \midrule
        \cite{lorenz2019}
         & \parbox[c]{\ww\linewidth}{\includegraphics[width=1.0\linewidth]{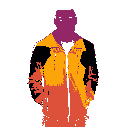}}
         & \parbox[c]{\ww\linewidth}{\includegraphics[width=1.0\linewidth]{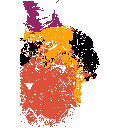}}
         & \parbox[c]{\ww\linewidth}{\includegraphics[width=1.0\linewidth]{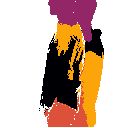}}
         & \parbox[c]{\ww\linewidth}{\includegraphics[width=1.0\linewidth]{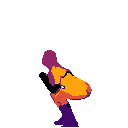}}
         & \parbox[c]{\ww\linewidth}{\includegraphics[width=1.0\linewidth]{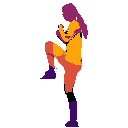}}
         & \parbox[c]{\ww\linewidth}{\includegraphics[width=1.0\linewidth]{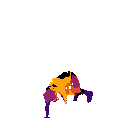}}
         & \parbox[c]{\ww\linewidth}{\includegraphics[width=1.0\linewidth]{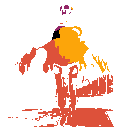}}
         & \parbox[c]{\ww\linewidth}{\includegraphics[width=1.0\linewidth]{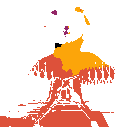}}
         & \parbox[c]{\ww\linewidth}{\includegraphics[width=1.0\linewidth]{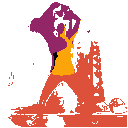}}
        \\
        \midrule
        Ours
         & \parbox[c]{\ww\linewidth}{\includegraphics[width=1.0\linewidth]{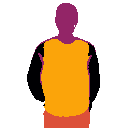}}
         & \parbox[c]{\ww\linewidth}{\includegraphics[width=1.0\linewidth]{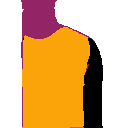}}
         & \parbox[c]{\ww\linewidth}{\includegraphics[width=1.0\linewidth]{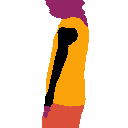}}
         & \parbox[c]{\ww\linewidth}{\includegraphics[width=1.0\linewidth]{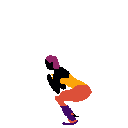}}
         & \parbox[c]{\ww\linewidth}{\includegraphics[width=1.0\linewidth]{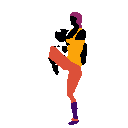}}
         & \parbox[c]{\ww\linewidth}{\includegraphics[width=1.0\linewidth]{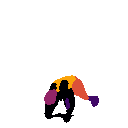}}
         & \parbox[c]{\ww\linewidth}{\includegraphics[width=1.0\linewidth]{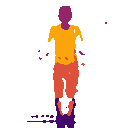}}
         & \parbox[c]{\ww\linewidth}{\includegraphics[width=1.0\linewidth]{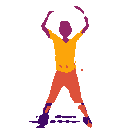}}
         & \parbox[c]{\ww\linewidth}{\includegraphics[width=1.0\linewidth]{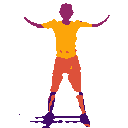}}
        \\
        \midrule
        GT
         & \parbox[c]{\ww\linewidth}{\includegraphics[width=1.0\linewidth]{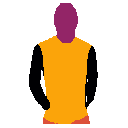}}
         & \parbox[c]{\ww\linewidth}{\includegraphics[width=1.0\linewidth]{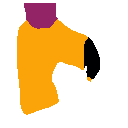}}
         & \parbox[c]{\ww\linewidth}{\includegraphics[width=1.0\linewidth]{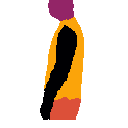}}
         & \parbox[c]{\ww\linewidth}{\includegraphics[width=1.0\linewidth]{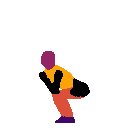}}
         & \parbox[c]{\ww\linewidth}{\includegraphics[width=1.0\linewidth]{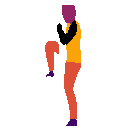}}
         & \parbox[c]{\ww\linewidth}{\includegraphics[width=1.0\linewidth]{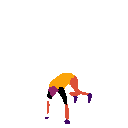}}
         & \parbox[c]{\ww\linewidth}{\includegraphics[width=1.0\linewidth]{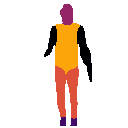}}
         & \parbox[c]{\ww\linewidth}{\includegraphics[width=1.0\linewidth]{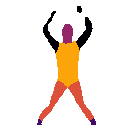}}
         & \parbox[c]{\ww\linewidth}{\includegraphics[width=1.0\linewidth]{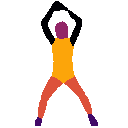}}
    \end{tabular}
    \caption{\textbf{Qualitative Comparison Against Keypoint Learning}.
        To obtain segmentation masks from keypoint baselines, we use an unsupervised postprocessing based on a conditional random field \cite{Krahenbuhl2012EfficientInferenceFully}.
        We do not apply any postprocessing on our results.
    }
    \label{fig:qualcompkeypoints}
\end{figure}

\subsection{Priors for Segmentations}

\begin{table*}[tb]
    \centering
    \begin{tabular}{lc|rrrrr|r}
        Dataset
         & Method
         & Arms
         & Feet
         & Head
         & Legs
         & Torso
         & Overall
        \\  \midrule
        \deepfashion{}
         & \cite{zhang2018unsupervised} + CRF
         & \fnum{0.194279}
         & \fnum{0.0}
         & \fnum{0.598378}
         & \fnum{0.292867}
         & \fnum{0.375825}
         & \fnum{0.29227}
        \\
        \deepfashion{}
         & \cite{jakab2018unsupervised} + CRF
         & \fnum{0.051952}
         & \fnum{0.0}
         & \fnum{0.117709}
         & \fnum{0.107532}
         & \fnum{0.244475}
         & \fnum{0.104334}
        \\
        \deepfashion{}
         & \cite{lorenz2019} + CRF
         & \fnum{0.214687}
         & \fnum{0.0}
         & \bfnum{0.605837}
         & \fnum{0.309113}
         & \fnum{0.321709}
         & \fnum{0.290269}
        \\
        \deepfashion{}
         & Ours
         & \bfnum{0.508}
         & \fnum{0.0}
         & \fnum{0.530}
         & \bfnum{0.500}
         & \bfnum{0.722}
         & \bfnum{0.45210}
        \\
        \midrule
        \exercise{}
         & \cite{zhang2018unsupervised} + CRF
         & \fnum{0.04279}
         & \bfnum{0.229842}
         & \fnum{0.0964388}
         & \fnum{0.432814}
         & \fnum{0.33541}
         & \fnum{0.227459}
        \\
        \exercise{}
         & \cite{jakab2018unsupervised} + CRF
         & \fnum{0.10138}
         & \fnum{0.189783}
         & \fnum{0.0}
         & \fnum{0.469242}
         & \fnum{0.356529}
         & \fnum{0.223387}
        \\
        \exercise{}
         & \cite{lorenz2019} + CRF
         & \fnum{0.211851}
         & \fnum{0.213396}
         & \bfnum{0.366471}
         & \bfnum{0.444965}
         & \fnum{0.441074}
         & \bfnum{0.335551}
        \\
        \exercise{}
         & Ours
         & \bfnum{0.252595}
         & \fnum{0.103587}
         & \fnum{0.340158}
         & \fnum{0.428429}
         & \bfnum{0.504409}
         & \fnum{0.325835}
        \\
        \midrule
        \pennaction{}
         & \cite{zhang2018unsupervised} + CRF
         & \fnum{0.0664641}
         & \fnum{0.0}
         & \bfnum{0.327259}
         & \fnum{0.379489}
         & \fnum{0.441596}
         & \fnum{0.242962}
        \\
        \pennaction{}
         & \cite{jakab2018unsupervised} + CRF
         & \fnum{0.0500828}
         & \bfnum{0.121661}
         & \fnum{0.}
         & \fnum{0.316491}
         & \fnum{0.454727}
         & \fnum{0.188592}
        \\
        \pennaction{}
         & \cite{lorenz2019} + CRF
         & \fnum{0.0376058}
         & \fnum{0.0}
         & \fnum{0.10532}
         & \fnum{0.311982}
         & \fnum{0.402122}
         & \fnum{0.171406}
        \\
        \pennaction{}
         & Ours
         & \bfnum{0.0937699}
         & \fnum{0.101009}
         & \fnum{0.236842}
         & \bfnum{0.370665}
         & \bfnum{0.484065}
         & \bfnum{0.25727}
    \end{tabular}
    \caption{\textbf{IOU Comparison Against Keypoint Learning}.
        To obtain segmentation masks from keypoint estimates, we use an
        unsupervised postprocessing based on a conditional random field
        \cite{Krahenbuhl2012EfficientInferenceFully}. See appendix for full details.
    }
    \label{tab:ioukeypoints}
\end{table*}

This section motivates suitable priors for the segmentation $S$.
We claim that part segmentations are locally smooth regions within the image,
meaning that long-range interactions between pixels are only possible through
local connectivity.
We illustrate this high-level idea in \figref{fig:metodsegmentationpriorsa}.
To achieve local smoothness within the image, we interpret $S$ as the output of a per-pixel classifier with probabilities $p_i(u, v), i = 1, \dots, N$.
We obtain $p_i(u, v)$ by a softmax normalization of the output of $\maskdecoder{}$, thus
\begin{align}
    \maskdecoder{} ~:
    ~ \pi \mapsto l, ~~~
    p_i(u, v)
    = \frac{
        \exp{
            \left(
            l_i(u, v)
            \right)
        }
    }{\sum_{i=1}^{N} \exp{
            \left(
            l_i(u, v)
            \right)
        }
    }.
\end{align}
In practice, $l$ can be seen as the logits of a classifier.
We now assume a Gaussian Markov Random Field prior for $l$, i.e.
$p(l) = \normal{0}{\nabla}$,
where $\nabla$ denotes the spatial gradient operator, which can be approximated using a finite-difference filter.
To efficiently train $\maskdecoder{}$, we use variational inference, meaning that we are looking for a suitable approximate posterior.
Using the mean-field approximation we can define
$
    q(l | x)
    = \prod_{i}^{\dim(l)} {
        q(l_i | x)
    }
    = \normal{\maskdecoder{}(\pi)}{I}.
$
Then, keeping $l$ close to the chosen prior in a $\kl$ sense simply results in regularizing the spatial gradient.
\begin{align}
    \kldiv{q}{p} = \sum_{i=1}^{N} \sum_{u, v} || \nabla_{(u, v)} l_i (u, v) ||^2
    \label{eq:klgmrf}
\end{align}
Unfortunately, this prior is not sufficient.
What is still missing is a prior that states that parts are mutually exclusive at every location, i.e. segmentations $S$ are categorical.
To enforce this, we have several options: using approximations of categorical distributions
\cite{Jang2016CategoricalReparameterizationGumbelSoftmax,Bengio2013EstimatingPropagatingGradients},
or add a regularizer that pushes the part segmentations towards a categorical solution, for instance by regularizing the entropy or cross-entropy, as shown in \figref{fig:metodsegmentationpriorsb}.
In practice, we found that entropy and cross-entropy regularization work best.
For simplicity, we restrict us to the entropy regularization.
\begin{align}
    \min~ \entropy{p} = \sum_{u, v} \sum_{i = 1}^{N} p_{i}(u, v) \log p_i(u, v)
\end{align}
Here, $(u, v)$ indicate spatial coordinate indices.
To summarize, we employ the following objective for $\maskdecoder{}$.
\begin{align}
    \maskdecoder{} ~:~ \min~ \loss{rec} + \lossweight{\text{GMRF}} \kldiv{q(l | x)}{p(l)} + \lossweight{\entropy{p}}\entropy{p}
\end{align}

\subsection{Part-based Image Generation}
\label{sec:part_based_image_generation}

\begin{figure}[tb]
    \def\ww{0.95}
    \begin{tabular}{l|c}
        Input
         & \parbox[c]{\ww\linewidth}{\includegraphics[width=1.0\linewidth]{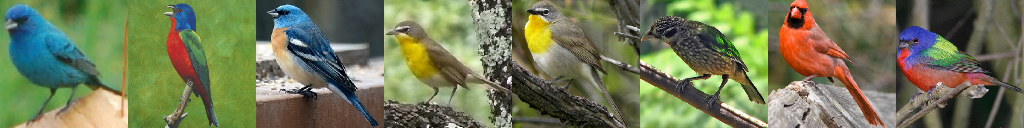}}
        \\
        \midrule
        \cite{zhang2018unsupervised}
         & \parbox[c]{\ww\linewidth}{\includegraphics[width=1.0\linewidth]{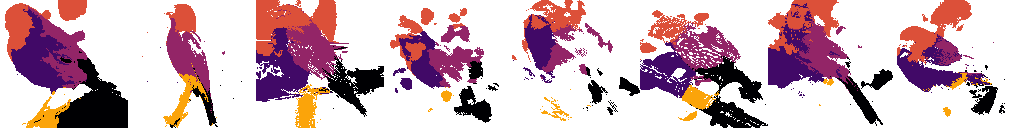}}
        \\
        \midrule
        \cite{jakab2018unsupervised}
         & \parbox[c]{\ww\linewidth}{\includegraphics[width=1.0\linewidth]{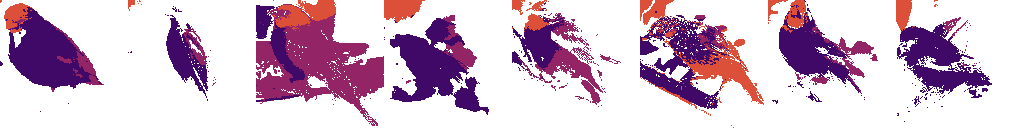}}
        \\
        \midrule
        \cite{lorenz2019}
         & \parbox[c]{\ww\linewidth}{\includegraphics[width=1.0\linewidth]{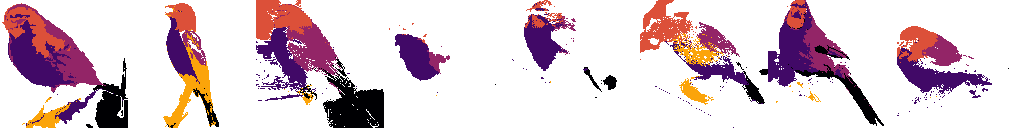}}
        \\
        \midrule
        Ours
         & \parbox[c]{\ww\linewidth}{\includegraphics[width=1.0\linewidth]{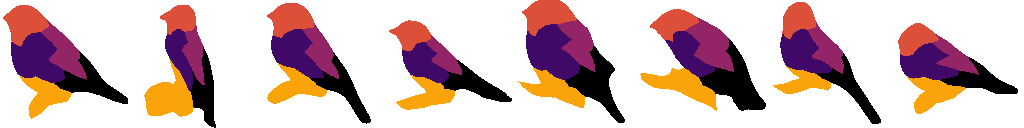}}
        \\
        \midrule
        GT
         & \parbox[c]{\ww\linewidth}{\includegraphics[width=1.0\linewidth]{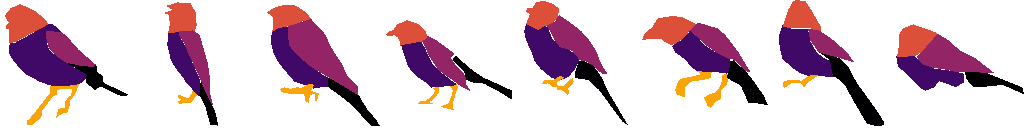}}
    \end{tabular}
    \hfill\null
    \caption{
        \textbf{Qualitative Results on \cub{}}.
        Despite the lack of multi-view training pairs, we are still able to learn a good part model using our proposed method.
    }
    \label{fig:celebaandcub}
\end{figure}

Having introduced all our chosen priors, it remains to specify the likelihood model $p(x | S)$ i.e. how we generate images $x$ from segmentations $S$.
Clearly, $S$ is not sufficient to explain the image because instance-specific appearance details are missing.
We therefore would like to build a part-based likelihood model that adds instance-specific part appearances $\alpha_i$.
We employ the following procedure, which is common practice in unsupervised keypoint learning \cite{lorenz2019,jakab2018unsupervised}, as shown in \figref{fig:methodcomplete}.
\begin{enumerate}
    \item Sample images $x_1$, $x_2$ from the dataset and infer their segmentations $S_1$ and $S_2$. As stated in \secref{sec:learning_appearance_independent_segmentations}, $x_1$ and $x_2$ are images of the same instance in different poses.
    \item
          Extract part based descriptors for appearance $\alphapart{2}{1}, \dots, \alphapart{2}{N}$ from $x_2$ by masking out each individual part using $S_2$ and mapping it into appearance space using $\encoderalpha{}$.
          The masking out operation is a simple hadamard product of each inferred part segmentation $x_{2, i} = S_{2, i} \odot x_2$ and can be interpreted as a part attention mechanism.
          We then obtain the part based descriptors using $\encoderalpha{}$\,:~
          $
              \alphapart{2}{i} = \encoderalpha{}(x_{2, i}).
          $
    \item We now have a set of vectors representing unlocalized part appearance descriptors and a spatial localization for those descriptors in terms of the segmentation $S$.
          To bring back spatial information for the appearance descriptors $\alphapart{2}{i}$ we calculate the expected appearance descriptor at each pixel which we term $S_\alpha$\,:~
          $
              S_\alpha (u, v) = \sum_{i=1}^{N} {\alphapart{2}{i} \cdot p_i(u, v)}
          $
    \item Finally, we reconstruct the image $x_1$ from $S_\alpha$ using a generator network $G$. More formally, this gives us the part-based image likelihood.
          \begin{align}
              p(x_1 | S_1, \alphapart{2}{1}, \dots, \alphapart{2}{N}) = \normal{G( S_\alpha )}{I}(x_1)
          \end{align}
\end{enumerate}
With this approach, we make the assumption that part appearances are constant across all poses $\pi$ for a specific instance.
Then, minimizing the negative log-likelihood gives the $\loss{rec}$ objective used in previous sections.
\begin{equation}
    \label{eq:lrec}
    \loss{rec} = - \log \left( \normal{G( S_\alpha )}{I}(x_1) \right)  = \norm{G( S_\alpha ) - x_1}^2
\end{equation}
In practice $G$ is a hour-glass style architecture
\cite{Newell2016StackedHourglassNetworks} and $\loss{rec}$ is implemented
through a perceptual loss \cite{johnsonPerceptualLossesRealTime2016}. See
supplementary for more details.

\section{Experiments}

\begin{table*}[tb]
    \centering
    \begin{tabular}{l|rrrrr|r}
        Method
         & Head
         & Chest
         & Wing
         & Tail
         & Feet
         & Overall
        \\
        \midrule
        \cite{zhang2018unsupervised} + CRF
         & \fnum{0.207192}
         & \fnum{0.320251}
         & \fnum{0.317655}
         & \fnum{0.364604}
         & \fnum{0.074242}
         & \fnum{0.256789}
        \\
        \cite{jakab2018unsupervised} + CRF
         & \fnum{0.0}
         & \fnum{0.393733}
         & \fnum{0.157655}
         & \fnum{0.189447}
         & \fnum{0.0}
         & \fnum{0.148167}
        \\
        \cite{lorenz2019} + CRF
         & \fnum{0.203219}
         & \fnum{0.477068}
         & \fnum{0.346727}
         & \fnum{0.431006}
         & \fnum{0.06768}
         & \fnum{0.30514}
        \\
        Ours
         & \bfnum{0.340228}
         & \bfnum{0.565433}
         & \bfnum{0.489071}
         & \bfnum{0.679168}
         & \bfnum{0.153745}
         & \bfnum{0.445529}
        \\
    \end{tabular}
    \caption{\textbf{IOU Comparison Against Keypoint Learning on Birds}.
        To obtain segmentation masks from keypoint baselines, we use an unsupervised postprocessing based on a conditional random field \cite{Krahenbuhl2012EfficientInferenceFully}. See appendix for details.
    }
    \label{tab:ioukeypointscub}
\end{table*}

\paragraph{Human Object Category}

We begin by evaluating our method on datasets of the human object category,
namely \deepfashion{} \cite{deepfashion1,deepfashion2}, \exercise{}
\cite{exercise1,exercise2} and \pennaction{} \cite{pennaction}.
\deepfashion{}
contains strong variations in viewpoints, poses and appearances but only a simple background.
\exercise{} has strong pose variation but only simple appearances and a simple background.
\pennaction{} introduces the additional challenge of background clutter.

We evaluate the performance of our method using the intersection-over-union (IOU) metric against a ground-truth part annotation.
We establish missing ground-truth annotation by using the supervised pretrained model from Densepose \cite{guler2018densepose} as a substitute oracle.
We calibrate our model on a held-out validation set to match the ground-truth as good as possible. Additional details can be found in the appendix.

We compare against recent work on unsupervised keypoint learning \cite{lorenz2019,jakab2018unsupervised,zhang2018unsupervised}.
To compare keypoint learning with segmentation learning, we apply a conditional random fields (CRF) \cite{Krahenbuhl2012EfficientInferenceFully} postprocessing.
This step is a standard technique to refine image segmentations \cite{Wang2015JointObjectPart,Tsogkas2015DeepLearningSemantic,Hung2019SCOPSSelfSupervisedCoPart}.
Note that we \emph{do not} apply any postprocessing on top of our proposed method.
Additional details can be found in the appendix.

Qualitative results of our method and keypoint learning baselines \cite{lorenz2019,jakab2018unsupervised,zhang2018unsupervised} are shown in \figref{fig:qualcompkeypoints}.
We observe that keypoint consistency is especially difficult to achieve when dealing with strong viewpoint variations,
for instance when switching between frontal and side poses on the \deepfashion{} and between push-up,
squatting position on the \exercise{} dataset.
The results on \pennaction{} suggest that background clutter is challenging for all methods, especially arm parts in downwards pointing poses.
Note that on some images with an extreme amount of part occlusions, even the supervised ground-truth model by \cite{guler2018densepose} fails to segment parts precisely (column 2, \figref{fig:qualcompkeypoints}).

Finally, we show quantitative results in terms of IOU in \tabref{tab:ioukeypoints}.
On \deepfashion{} and \pennaction{} our method outperforms other methods by a consistent margin in terms of IOU.
On \exercise{}, the method is on par with the state-of-the art keypoint model \cite{lorenz2019} paired with CRF postprocessing.
The quantitative results validate our observation for all the datasets that our method is able to discover
semantically consistent parts across instances in form of segmentations.

\begin{figure}[tb]
    \centering
    \null\hfill
    \begin{subfigure}[t]{0.3\textwidth}
        \centering
        \parbox[c]{1\linewidth}{\includegraphics[width=1\linewidth]{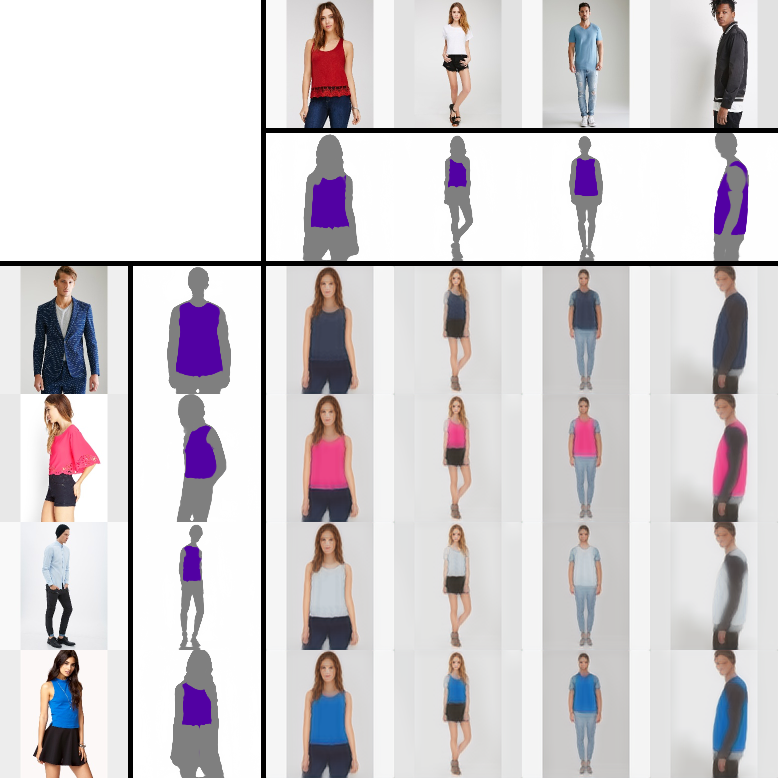}}
        \caption{
            Swapping only chest appearance.
        }
    \end{subfigure}
    \hfill
    \begin{subfigure}[t]{0.3\textwidth}
        \centering
        \parbox[c]{1\linewidth}{\includegraphics[width=1\linewidth]{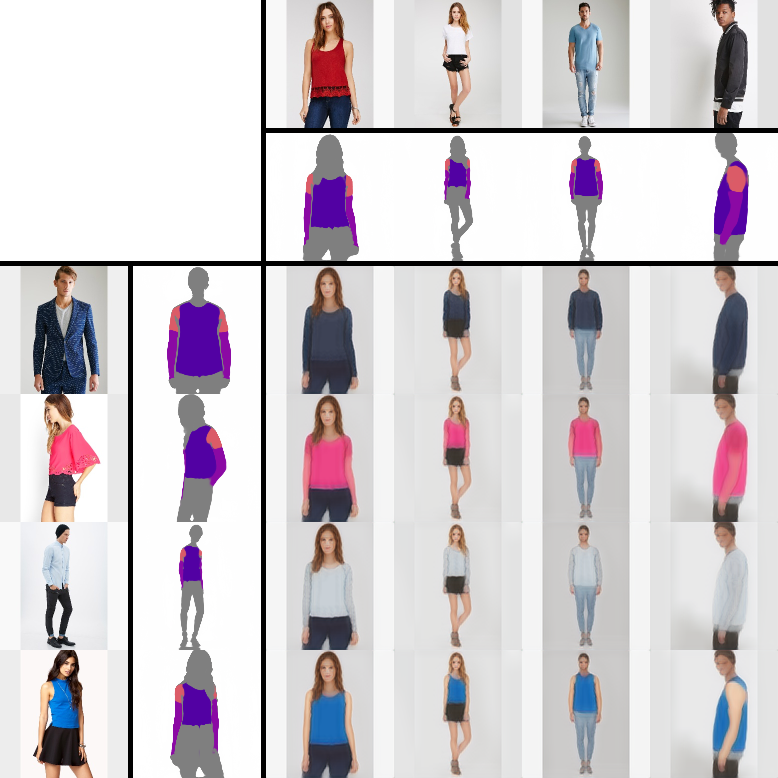}}
        \caption{
            Swapping chest and arm appearance.
        }
    \end{subfigure}
    \hfill
    \begin{subfigure}[t]{0.3\textwidth}
        \centering
        \parbox[c]{1\linewidth}{\includegraphics[width=1\linewidth]{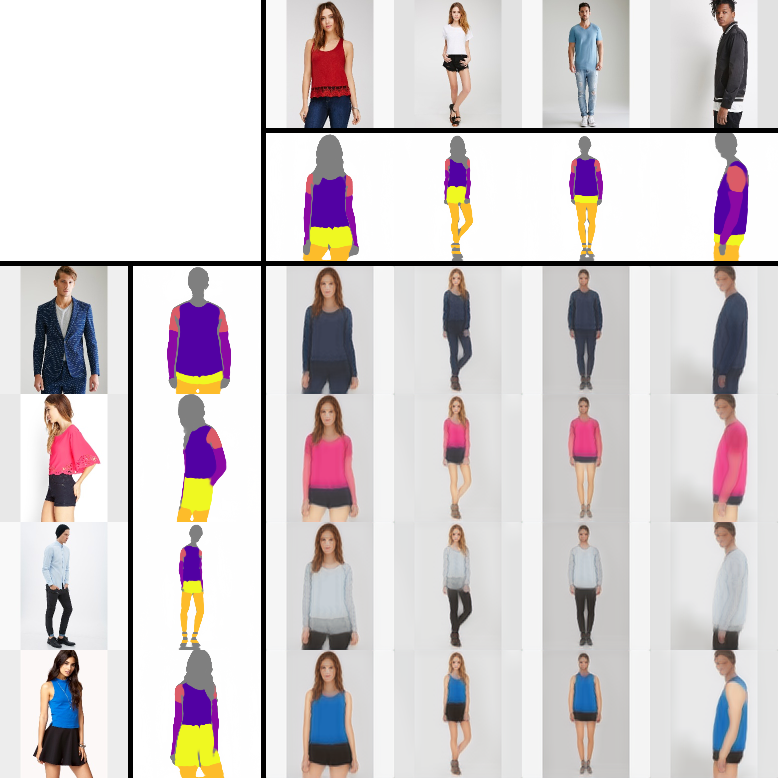}}
        \caption{
            Swapping chest, arm, hip and leg appearance.
        }
    \end{subfigure}
    \hfill\null
    \caption{
        \textbf{Part-based Appearance Transfer}.
        Parts which are swapped are highlighted in color (active), parts which remain constant are gray (inactive).
        (a): we transfer appearance of torso parts.
        (b): we transfer appearance of torso and arm parts.
        (c): we transfer appearance of torso, arm and leg parts.
        The transfer succeeds despite strong occlusions and viewpoint variations.
    }
    \label{fig:transfer}
\end{figure}

\paragraph{Other Object Categories}

We qualitatively analyze our method on the bird object category in \figref{fig:celebaandcub}.
Note that \cub{} is a single image dataset, which requires us to use
artificial thin-plate-spline transformations (TPS) as an approximation to
multi-view pose variations.
This approximation is identical to those used in
\cite{Thewlis2017UnsupervisedLearningObject,jakab2018unsupervised,lorenz2019}.
We observe that our part discovery method learns local parts and is also able to find
appropriate scales for parts for smaller sized birds.
To evaluate our method quantitatively,
we created a small dataset of bird part annotations as ground-truth information and
evaluate against \cite{zhang2018unsupervised,jakab2018unsupervised,lorenz2019} in terms of IOU in \tabref{tab:ioukeypointscub}.
The results suggest that our approach can be scaled to other object categories.
\begin{table}[tb]
    \centering
    \begin{tabular}{l|cccc}
        $\alpha$
         & PCK@\SI{2.5}{\percent}
         & PCK@\SI{5}{\percent}
         & PCK@\SI{10}{\percent}
        \\
        \hline
        VU-Net \cite{Esser2018VariationalUNetConditionala}
         & 31.64
         & 54.90
         & 80.83
        \\
        Lorenz et al. \cite{lorenz2019}
         & 14.50
         & 37.50
         & 69.63
        \\
        Ours
         & \textbf{41.56}
         & \textbf{65.76}
         & \textbf{83.12}
        \\
    \end{tabular}
    \caption{\textbf{Evaluating Shape Consistency}.
        Percentage of Correct Keypoints (PCK) for pose estimation on shape/appearance swapped generations for supervised and unsupervised methods.
        $\alpha$ is pixel distance divided by image diagonal.
        Note that \cite{Esser2018VariationalUNetConditionala} serves as upper bound, as it uses the groundtruth shape estimates.
    }
    \label{tab:PCK}
\end{table}

\subsection{Part-based Appearance Transfer}

We explore the capabilities of part-based appearance transfer between instances in \figref{fig:transfer}.
Parts which are transferred are displayed in color (active), parts which are not transferred are displayed in gray (inactive).
The transfer succeeds despite strong occlusions and pose variations.
In the most extreme case, occluded appearances can be inferred from partial observations,
for instance when transferring from half-body images to full-body images
or from frontal to side-ways poses.
Note that we do not use any adversarial training, which causes our generated images to look rather smooth and untextured in comparison to state-of-the art image synthesis.

Following \cite{lorenz2019},
we evaluate the resulting pose consistency when transferring parts between instances by
calculating the percentage of correct keypoints after swapping the appearance.
The results in \tabref{tab:PCK} show that our method performs significantly better than \cite{lorenz2019}
and even outperforms the supervised baseline VU-Net \cite{Esser2018VariationalUNetConditionala} by a small margin.

Due to space constraints, we refer the reader to the supplementary materials
regarding an ablation study.

\section{Conclusion}

We have shown that we can build a generative model for part segmentations by a suitable combination of priors.
Since the method is generative, it allows learning part segmentations without explicit supervision.
Experiments demonstrate the benefits of this approach over models
which obtain part masks through keypoints. Overall, this work shows that disentanglement serves as
a powerful substitute for supervision and, when combined with appropriate
priors, allows to directly discover part segmentations. This is in
contrast to most previous works on unsupervised learning, which consider
unsupervised learning merely as a pre-training step to be followed by
supervised training to obtain the final result.

\section{Acknowledgements}

This work has been supported in part by the BW Stiftung project ``MULT!nano'', the German Research Foundation (DFG) project 421703927, and the German federal ministry BMWi within the project ``KI Absicherung''.

%% file: figures/svg/method2.pdf_tex
\begingroup%
  \makeatletter%
  \providecommand\color[2][]{%
    \errmessage{(Inkscape) Color is used for the text in Inkscape, but the package 'color.sty' is not loaded}%
    \renewcommand\color[2][]{}%
  }%
  \providecommand\transparent[1]{%
    \errmessage{(Inkscape) Transparency is used (non-zero) for the text in Inkscape, but the package 'transparent.sty' is not loaded}%
    \renewcommand\transparent[1]{}%
  }%
  \providecommand\rotatebox[2]{#2}%
  \newcommand*\fsize{\dimexpr\f@size pt\relax}%
  \newcommand*\lineheight[1]{\fontsize{\fsize}{#1\fsize}\selectfont}%
  \ifx\svgwidth\undefined%
    \setlength{\unitlength}{677.12983379bp}%
    \ifx\svgscale\undefined%
      \relax%
    \else%
      \setlength{\unitlength}{\unitlength * \real{\svgscale}}%
    \fi%
  \else%
    \setlength{\unitlength}{\svgwidth}%
  \fi%
  \global\let\svgwidth\undefined%
  \global\let\svgscale\undefined%
  \makeatother%
  \begin{picture}(1,0.44171441)%
    \lineheight{1}%
    \setlength\tabcolsep{0pt}%
    \put(0,0){\includegraphics[width=\unitlength,page=1]{method2.pdf}}%
    \put(0.14209089,0.3731157){\color[rgb]{0,0,0}\makebox(0,0)[lt]{\lineheight{1.25}\smash{\begin{tabular}[t]{l}$x_2$\end{tabular}}}}%
    \put(0.14209088,0.20628642){\color[rgb]{0,0,0}\makebox(0,0)[lt]{\lineheight{1.25}\smash{\begin{tabular}[t]{l}$x_1$\end{tabular}}}}%
    \put(0.43102773,0.12587742){\color[rgb]{0,0,0}\makebox(0,0)[t]{\lineheight{1.25}\smash{\begin{tabular}[t]{c}$\pi \sim \normal{\mu(x_1)}{\Sigma(x_1)}$\end{tabular}}}}%
    \put(0,0){\includegraphics[width=\unitlength,page=2]{method2.pdf}}%
    \put(0.38534517,0.37403133){\color[rgb]{0,0,0}\makebox(0,0)[lt]{\lineheight{1.25}\smash{\begin{tabular}[t]{l}$\alpha$\end{tabular}}}}%
    \put(0.38213192,0.20636751){\color[rgb]{0,0,0}\makebox(0,0)[lt]{\lineheight{1.25}\smash{\begin{tabular}[t]{l}$\pi$\end{tabular}}}}%
    \put(0,0){\includegraphics[width=\unitlength,page=3]{method2.pdf}}%
    \put(0.18333015,0.22428156){\color[rgb]{0,0,0}\makebox(0,0)[lt]{\begin{minipage}{0.20251394\unitlength}\raggedright $\encoderpi{}$\\ \end{minipage}}}%
    \put(0,0){\includegraphics[width=\unitlength,page=4]{method2.pdf}}%
    \put(0.18067233,0.39171041){\color[rgb]{0,0,0}\makebox(0,0)[lt]{\begin{minipage}{0.20251394\unitlength}\raggedright $\encoderalpha{}$\\ \end{minipage}}}%
    \put(0.74095994,0.05736407){\color[rgb]{0,0,0}\makebox(0,0)[lt]{\lineheight{1.25}\smash{\begin{tabular}[t]{l}$S$\end{tabular}}}}%
    \put(0,0){\includegraphics[width=\unitlength,page=5]{method2.pdf}}%
    \put(0.53952299,0.07781111){\color[rgb]{0,0,0}\makebox(0,0)[lt]{\begin{minipage}{0.16671044\unitlength}\raggedleft $\maskdecoder{}$\end{minipage}}}%
    \put(0,0){\includegraphics[width=\unitlength,page=6]{method2.pdf}}%
    \put(0.6109054,0.40433902){\color[rgb]{0,0,0}\makebox(0,0)[t]{\lineheight{1.25}\smash{\begin{tabular}[t]{c}$\pi \sim p(\pi), \alpha \sim p(\alpha)$\end{tabular}}}}%
    \put(0,0){\includegraphics[width=\unitlength,page=7]{method2.pdf}}%
    \put(0.76265466,0.30605634){\color[rgb]{0,0,0}\makebox(0,0)[lt]{\begin{minipage}{0.20251394\unitlength}\raggedright $T$\\ \end{minipage}}}%
    \put(0.8622042,0.28810309){\color[rgb]{0,0,0}\makebox(0,0)[lt]{\lineheight{1.25}\smash{\begin{tabular}[t]{l}$I_{T}(\pi,\alpha)$\end{tabular}}}}%
    \put(0,0){\includegraphics[width=\unitlength,page=8]{method2.pdf}}%
    \put(0.6109054,0.17075752){\color[rgb]{0,0,0}\makebox(0,0)[t]{\lineheight{1.25}\smash{\begin{tabular}[t]{c}$(\pi, \alpha) \sim p(\pi, \alpha)$\end{tabular}}}}%
    \put(0,0){\includegraphics[width=\unitlength,page=9]{method2.pdf}}%
  \end{picture}%
\endgroup%

%% file: figures/svg/method1.pdf_tex
\begingroup%
  \makeatletter%
  \providecommand\color[2][]{%
    \errmessage{(Inkscape) Color is used for the text in Inkscape, but the package 'color.sty' is not loaded}%
    \renewcommand\color[2][]{}%
  }%
  \providecommand\transparent[1]{%
    \errmessage{(Inkscape) Transparency is used (non-zero) for the text in Inkscape, but the package 'transparent.sty' is not loaded}%
    \renewcommand\transparent[1]{}%
  }%
  \providecommand\rotatebox[2]{#2}%
  \newcommand*\fsize{\dimexpr\f@size pt\relax}%
  \newcommand*\lineheight[1]{\fontsize{\fsize}{#1\fsize}\selectfont}%
  \ifx\svgwidth\undefined%
    \setlength{\unitlength}{783.24932909bp}%
    \ifx\svgscale\undefined%
      \relax%
    \else%
      \setlength{\unitlength}{\unitlength * \real{\svgscale}}%
    \fi%
  \else%
    \setlength{\unitlength}{\svgwidth}%
  \fi%
  \global\let\svgwidth\undefined%
  \global\let\svgscale\undefined%
  \makeatother%
  \begin{picture}(1,0.41363491)%
    \lineheight{1}%
    \setlength\tabcolsep{0pt}%
    \put(0,0){\includegraphics[width=\unitlength,page=1]{method1.pdf}}%
    \put(0.12283953,0.19219944){\color[rgb]{0,0,0}\makebox(0,0)[lt]{\lineheight{1.25}\smash{\begin{tabular}[t]{l}$x_2$\end{tabular}}}}%
    \put(0.12283951,0.35438909){\color[rgb]{0,0,0}\makebox(0,0)[lt]{\lineheight{1.25}\smash{\begin{tabular}[t]{l}$x_1$\end{tabular}}}}%
    \put(0.2625112,0.27471856){\color[rgb]{0,0,0}\makebox(0,0)[t]{\lineheight{1.25}\smash{\begin{tabular}[t]{c}$\pi_i \sim \normal{\mu(x_i)}{\Sigma(x_i)}$\end{tabular}}}}%
    \put(0,0){\includegraphics[width=\unitlength,page=2]{method1.pdf}}%
    \put(0.46575499,0.06369669){\color[rgb]{0,0,0}\makebox(0,0)[lt]{\begin{minipage}{0.17507609\unitlength}\raggedright $\encoderalpha{}$\\ \end{minipage}}}%
    \put(0.59544499,0.00361322){\color[rgb]{0,0,0}\makebox(0,0)[lt]{\lineheight{1.25}\smash{\begin{tabular}[t]{l}$\alpha^2_1$\end{tabular}}}}%
    \put(0,0){\includegraphics[width=\unitlength,page=3]{method1.pdf}}%
    \put(0.25307031,0.4019947){\color[rgb]{0,0,0}\makebox(0,0)[lt]{\lineheight{1.25}\smash{\begin{tabular}[t]{l}$\pi_1$\end{tabular}}}}%
    \put(0,0){\includegraphics[width=\unitlength,page=4]{method1.pdf}}%
    \put(0.15849144,0.3699461){\color[rgb]{0,0,0}\makebox(0,0)[lt]{\begin{minipage}{0.17507609\unitlength}\raggedright $\encoderpi{}$\\ \end{minipage}}}%
    \put(0,0){\includegraphics[width=\unitlength,page=5]{method1.pdf}}%
    \put(0.25450659,0.14318099){\color[rgb]{0,0,0}\makebox(0,0)[lt]{\lineheight{1.25}\smash{\begin{tabular}[t]{l}$\pi_2$\end{tabular}}}}%
    \put(0,0){\includegraphics[width=\unitlength,page=6]{method1.pdf}}%
    \put(0.37689217,0.19252349){\color[rgb]{0,0,0}\makebox(0,0)[lt]{\lineheight{1.25}\smash{\begin{tabular}[t]{l}$S_2$\end{tabular}}}}%
    \put(0,0){\includegraphics[width=\unitlength,page=7]{method1.pdf}}%
    \put(0.21656284,0.21020022){\color[rgb]{0,0,0}\makebox(0,0)[lt]{\begin{minipage}{0.14412347\unitlength}\raggedleft $\maskdecoder{}$\end{minipage}}}%
    \put(0.15849144,0.20825526){\color[rgb]{0,0,0}\makebox(0,0)[lt]{\begin{minipage}{0.17507609\unitlength}\raggedright $\encoderpi{}$\\ \end{minipage}}}%
    \put(0.37771292,0.35421433){\color[rgb]{0,0,0}\makebox(0,0)[lt]{\lineheight{1.25}\smash{\begin{tabular}[t]{l}$S_1$\end{tabular}}}}%
    \put(0,0){\includegraphics[width=\unitlength,page=8]{method1.pdf}}%
    \put(0.21656284,0.37189108){\color[rgb]{0,0,0}\makebox(0,0)[lt]{\begin{minipage}{0.14412347\unitlength}\raggedleft $\maskdecoder{}$\end{minipage}}}%
    \put(0,0){\includegraphics[width=\unitlength,page=9]{method1.pdf}}%
    \put(0.34936427,0.04833864){\color[rgb]{0,0,0}\makebox(0,0)[lt]{\lineheight{1.25}\smash{\begin{tabular}[t]{l}$S_2 \odot x_2$\end{tabular}}}}%
    \put(0.66602207,0.00361322){\color[rgb]{0,0,0}\makebox(0,0)[lt]{\lineheight{1.25}\smash{\begin{tabular}[t]{l}$\alpha^2_n$\end{tabular}}}}%
    \put(0,0){\includegraphics[width=\unitlength,page=10]{method1.pdf}}%
    \put(0.60680229,0.0487733){\color[rgb]{0,0,0}\makebox(0,0)[lt]{\lineheight{1.25}\smash{\begin{tabular}[t]{l}$\dots$\end{tabular}}}}%
    \put(0,0){\includegraphics[width=\unitlength,page=11]{method1.pdf}}%
    \put(0.77210025,0.32688162){\color[rgb]{0,0,0}\makebox(0,0)[lt]{\lineheight{1.25}\smash{\begin{tabular}[t]{l}$G$\end{tabular}}}}%
    \put(0.86432754,0.35748609){\color[rgb]{0,0,0}\makebox(0,0)[lt]{\lineheight{1.25}\smash{\begin{tabular}[t]{l}$\hat{x}_1$\end{tabular}}}}%
    \put(0,0){\includegraphics[width=\unitlength,page=12]{method1.pdf}}%
    \put(0.55228478,0.33812018){\color[rgb]{0,0,0}\makebox(0,0)[lt]{\lineheight{1.25}\smash{\begin{tabular}[t]{l}$\alpha^2_1, \dots, \alpha^2_n$\end{tabular}}}}%
    \put(0.60164202,0.36120817){\color[rgb]{0,0,0}\makebox(0,0)[lt]{\lineheight{1.25}\smash{\begin{tabular}[t]{l}$S_1$\end{tabular}}}}%
    \put(0,0){\includegraphics[width=\unitlength,page=13]{method1.pdf}}%
  \end{picture}%
\endgroup%

%% file: tex_files/appendix.tex
\section{Appendix}

\begin{table}
    \centering
    \begin{tabular}{l|c|c}
        Operation         & Input shape      & Output shape          \\
        \hline
        Input             & $(1, 1, 256)$    & $(1, 1, 256)$         \\
        $1\times1$ Conv2D & $(1, 1, 256)$    & $(1, 1, 4096)$        \\
        Conv2D            & $(4, 4, 256)$    & $(4, 4, 256)$         \\
        residual block    & $(4, 4, 256)$    & $(4, 4, 256)$         \\
        residual block    & $(4, 4, 256)$    & $(4, 4, 256)$         \\
        Upsample          & $(4, 4, 256)$    & $(8, 8, 128)$         \\
        residual block    & $(8, 8, 128)$    & $(8, 8, 128)$         \\
        Upsample          & $(8, 8, 128)$    & $(16, 16, 128)$       \\
        residual block    & $(16, 16, 128)$  & $(16, 16, 128)$       \\
        Upsample          & $(16, 16, 32)$   & $(32, 32, 32)$        \\
        residual block    & $(32, 32, 32)$   & $(32, 32, 32)$        \\
        Upsample          & $(32, 32, 32)$   & $(64, 64, 32)$        \\
        residual block    & $(64, 64, 32)$   & $(64, 64, 32)$        \\
        Upsample          & $(64, 64, 32)$   & $(128, 128, 16)$      \\
        residual block    & $(128, 128, 16)$ & $(128, 128, 16)$      \\
        residual block    & $(128, 128, 16)$ & $(128, 128, 16)$      \\
        Conv2D            & $(128, 128, 16)$ & $(128, 128, n_{out})$
    \end{tabular}
    \caption{$D_m$ architecture. $n_{out}$ was set to the maximum number of parts discovered. All \textit{Conv2D} layer are coord-convs \cite{liuIntriguingFailingConvolutional2018}.}
    \label{tab:maskdecoder}
\end{table}

\begin{table}
    \centering
    \begin{tabular}{l|c|c}
        Operation                 & Input shape      & Output shape      \\
        \hline
        Input                     & $(128, 128, 3)$  & $(128, 128, 3)$   \\
        Conv2D                    & $(128, 128, 3)$  & $(128, 128, 16)$  \\
        residual block            & $(128, 128, 16)$ & $(128, 128, 16)$  \\
        Conv2D with stride 2      & $(128, 128, 16)$ & $(64, 64, 32)$    \\
        residual block            & $(64, 64, 32)$   & $(64, 64, 32)$    \\
        Conv2D with stride 2      & $(64, 64, 32)$   & $(32, 32, 64)$    \\
        residual block            & $(32, 32, 64)$   & $(32, 32, 64)$    \\
        Conv2D with stride 2      & $(32, 32, 64)$   & $(16, 16, 128)$   \\
        residual block            & $(16, 16, 128)$  & $(16, 16, 128)$   \\
        Conv2D with stride 2      & $(16, 16, 128)$  & $(8, 8, 128)$     \\
        residual block            & $(8, 8, 128)$    & $(8, 8, 128)$     \\
        Conv2D with stride 2      & $(8, 8, 128)$    & $(4, 4, 256)$     \\
        residual block            & $(4, 4, 256)$    & $(4, 4, 256)$     \\
        $4 \times$ residual block & $(4, 4, 256)$    & $(4, 4, 256)$     \\
        mean pooling              & $(4, 4, 256)$    & $(1, 1, 256)$     \\
        $1\times1$ Conv2D         & $(1, 1, 256)$    & $(1, 1, n_{out})$
    \end{tabular}
    \caption{
        $E_\alpha$ and $E_\beta$ architectures. For $E_\beta$, we use coord-convs \cite{liuIntriguingFailingConvolutional2018}.
        This means that we additionally concatenate the spatial coordinates to the feature maps before convolution. $n_{out}$ is 256 for $E_\beta$ and 64 for $E_\alpha$.
    }
    \label{tab:encoders}
\end{table}

\begin{table}
    \centering
    \begin{tabular}{l|c|c}
        Operation            & Input shape             & Output shape     \\
        \hline
        Input                & $(128, 128, 89)$        & $(128, 128, 89)$ \\
        Conv2D               & $(128, 128, 89)$        & $(128, 128, 32)$ \\
        residual block       & $(128, 128, 32)$        & $(128, 128, 32)$ \\
        Conv2D with stride 2 & $(128, 128, 32)$        & $(64, 64, 64)$   \\
        residual block (a)   & $(64, 64, 64)$          & $(64, 64, 64)$   \\
        residual block       & $(64, 64, 64)$          & $(64, 64, 64)$   \\
        \begin{tabular}{@{}c@{}} residual block \\ with skip from (a) \end{tabular}
                             & $2 \times (64, 64, 64)$ & $(64, 64, 64)$   \\
        Upsample bilinear    & $(128, 128, 32)$        & $(128, 128, 32)$ \\
        residual block       & $(128, 128, 32)$        & $(128, 128, 32)$ \\
        Conv 2D              & $(128, 128, 32)$        & $(128, 128, 3)$  \\
    \end{tabular}
    \caption{$G$ architecture, resembling a shallow hourglass network. Incoming skip connections are first passed through the activation function, then convolved with a $1 \times 1$-conv2D and then concatenated to the input from the upsampling stage.}
    \label{tab:g}
\end{table}

\subsection{Implementation Details}

The neural networks used in our model are provided in \tabref{tab:maskdecoder}, \tabref{tab:encoders} and \tabref{tab:g}.
All \texttt{Conv2D} layers use filters of size $3 \times 3$. A residual block with input $x$ and output $y$ is defined as follows:
\begin{align}
    a(x) & = \texttt{leaky\_relu}(x)    \\
    y    & = \texttt{conv2D}(a(x)) + x.
\end{align}
A residual block with input $x$, incoming skip-connections $i$ and output $y$ is defined as follows:
\begin{align}
    a(x) & = \texttt{leaky\_relu}(x) \\
    c    & = [a(x),
    \texttt{$1\times1$conv2D}( a(\mathrm{i}) )
    ]                                \\
    y    & = \texttt{conv2D}
    \left( c
    \right) + x.
\end{align}

For all experiments, the same architectures were used.
All networks were initialized using the standard initialization introduced by \cite{heDelvingDeepRectifiers2015}.
The input images were resized to a shape of $128 \times 128$.
The \emph{Number of parts} - parameter $N$ is set to an arbitrary number that is sufficiently high and provides and upper bound on the discovered parts.
We choose 25 for all experiments, without loss of generality.
We apply no data augmentation other than horizontal flipping.
We train our model with batch size $4$ for $\num{100000}$ steps using
the Adam optimizer \cite{Kingma2014AdamMethodStochastic} with an initial learning rate of $2 \cdot 10^{-4}$.
$I_T$ is calculated with an exponential moving average with a decay of $0.99$.
The value of $\lossweight{\text{GMRF}}$ is set to $1.0 \cdot 10^{-3}$.
The value of $\lossweight{\entropy{p}}$ is set to $0.06 \cdot 10^{-3}$ at the beginning of the training and linearly increased between $\num{30000}$ and $\num{50000}$ steps to $\num{0.06}$.
The dimensionalities of latent variables are: $\dim(\alpha) = 128$ and $\dim (\pi) = 64$.

We provide details on the datasets that were used in our experiments:
\begin{description}
    \item[Deepfashion] We use the train and test split provided by \cite{deepfashion1,deepfashion2},
          consisting of $\num{31802}$ for training and $\num{984}$ for testing.
    \item[Pennaction] We use a subset of the train and test split provided by \cite{pennaction},
          consisting of $\num{1648}$ for training and $\num{1689}$ for testing.
    \item[CUB] We use the train and test split provided by \cite{lorenz2019},
          consisting of $\num{4736}$ images for training and $\num{4631}$ for testing.
\end{description}

\subsection{Keypoint Baselines, Postprocessing and Calibration Step}

\begin{figure}[tb]
    \centering
    \footnotesize
    \def\svgwidth{\textwidth}
    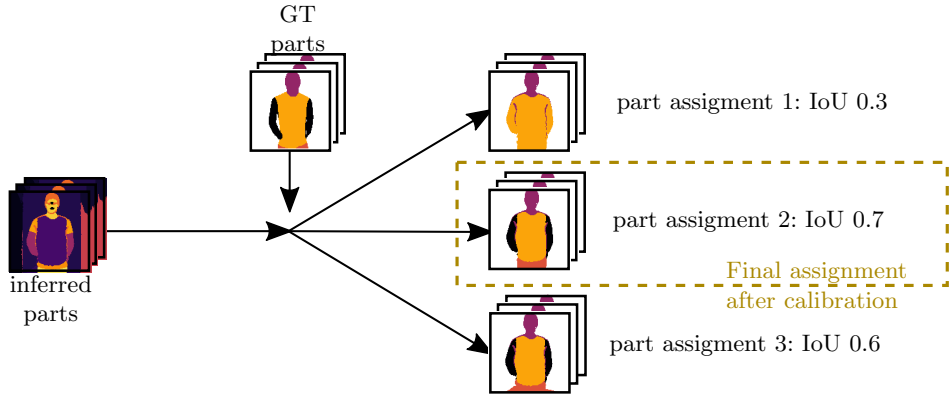
    \caption{
        \textbf{Calibration Step}.
        Given the set of inferred parts, the calibration step calculates the IoU with each possible assignment of individual ground-truth parts.
        Afterwards, the assignment with the best IoU score is used for evaluation.
    }
    \label{fig:calibration_step}
\end{figure}

\begin{table*}
    \centering
    \begin{tabular}{lc|c|rrrrrr}
        Dataset        & Method                       & $\sigma$      & Arms                & Feet               & Head                 & Legs               & Torso               & Overall             \\ \midrule
        \deepfashion{} & \cite{zhang2018unsupervised} & $\fnum{0.01}$ & $\fnum{0.00629344}$ & $\fnum{0}$         & $\fnum{0.0144026}$   & $\fnum{0.0}$       & $\fnum{0.00784737}$ & $\fnum{0.00570867}$ \\
        \deepfashion{} & \cite{zhang2018unsupervised} & $\fnum{0.05}$ & $\ufnum{0.231726}$  & $\fnum{0}$         & $\fnum{0.578507}$    & $\fnum{0.295842}$  & $\fnum{0.30743}$    & $\fnum{0.282701}$   \\
        \deepfashion{} & \cite{zhang2018unsupervised} & $\fnum{0.1}$  & $\fnum{0.194279}$   & $\fnum{0}$         & $\ufnum{0.598378}$   & $\fnum{0.292867}$  & $\ufnum{0.375825}$  & $\fnum{0.29227}$    \\
        \hline
        \deepfashion{} & \cite{jakab2018unsupervised} & $\fnum{0.01}$ & $\fnum{0.00199367}$ & $\fnum{0}$         & $\fnum{0.000525233}$ & $\fnum{0.0}$       & $\fnum{0.00439446}$ & $\fnum{0.00138267}$ \\
        \deepfashion{} & \cite{jakab2018unsupervised} & $\fnum{0.05}$ & $\fnum{0.00215036}$ & $\fnum{0}$         & $\fnum{0.0423404}$   & $\fnum{0.0558477}$ & $\fnum{0.0876187}$  & $\fnum{0.0375914}$  \\
        \deepfashion{} & \cite{jakab2018unsupervised} & $\fnum{0.1}$  & $\fnum{0.051952}$   & $\fnum{0}$         & $\fnum{0.117709}$    & $\fnum{0.107532}$  & $\fnum{0.244475}$   & $\fnum{0.104334}$   \\
        \hline
        \deepfashion{} & \cite{lorenz2019}            & $\fnum{0.01}$ & $\fnum{0.00201808}$ & $\fnum{0}$         & $\fnum{0.0085051}$   & $\fnum{0.0}$       & $\fnum{0.0135235}$  & $\fnum{0.00480934}$ \\
        \deepfashion{} & \cite{lorenz2019}            & $\fnum{0.05}$ & $\fnum{0.106806}$   & $\fnum{0}$         & $\fnum{0.272331}$    & $\fnum{0.180952}$  & $\fnum{0.318051}$   & $\fnum{0.175628}$   \\
        \deepfashion{} & \cite{lorenz2019}            & $\fnum{0.1}$  & $\fnum{0.214687}$   & $\fnum{0}$         & $\bfnum{0.605837}$   & $\ufnum{0.309113}$ & $\fnum{0.321709}$   & $\ufnum{0.290269}$  \\
        \hline
        \deepfashion{} & Ours                         & -             & \bfnum{0.508}       & \fnum{0.0}         & \fnum{0.530}         & \bfnum{0.500}      & \bfnum{0.722}       & \bfnum{0.45210}     \\
        \midrule
        \exercise{}    & \cite{zhang2018unsupervised} & $\fnum{0.01}$ & $\fnum{0.00389304}$ & $\fnum{0.0862982}$ & $\fnum{0.0161744}$   & $\fnum{0.0322612}$ & $\fnum{0.201313}$   & $\fnum{0.067988}$   \\
        \exercise{}    & \cite{zhang2018unsupervised} & $\fnum{0.05}$ & $\fnum{0.0406129}$  & $\bfnum{0.225912}$ & $\fnum{0.102956}$    & $\fnum{0.427865}$  & $\fnum{0.332834}$   & $\fnum{0.226036}$   \\
        \exercise{}    & \cite{zhang2018unsupervised} & $\fnum{0.1}$  & $\fnum{0.0234757}$  & $\fnum{0.0652997}$ & $\fnum{0.0205852}$   & $\fnum{0.250434}$  & $\fnum{0.240598}$   & $\fnum{0.120079}$   \\
        \hline
        \exercise{}    & \cite{jakab2018unsupervised} & $\fnum{0.01}$ & $\fnum{0.0}$        & $\fnum{0.0546221}$ & $\fnum{0.0}$         & $\fnum{0.131787}$  & $\fnum{0.274826}$   & $\fnum{0.092247}$   \\
        \exercise{}    & \cite{jakab2018unsupervised} & $\fnum{0.05}$ & $\fnum{0.0907609}$  & $\fnum{0.174806}$  & $\fnum{0.102956}$    & $\ufnum{0.462166}$ & $\fnum{0.380064}$   & $\fnum{0.22156}$    \\
        \exercise{}    & \cite{jakab2018unsupervised} & $\fnum{0.1}$  & $\fnum{0.097578}$   & $\fnum{0.192397}$  & $\fnum{0.0}$         & $\bfnum{0.464227}$ & $\fnum{0.350028}$   & $\fnum{0.220846}$   \\
        \hline
        \exercise{}    & \cite{lorenz2019}            & $\fnum{0.01}$ & $\fnum{0.133528}$   & $\fnum{0.146215}$  & $\bfnum{0.395451}$   & $\fnum{0.332499}$  & $\fnum{0.42839}$    & $\fnum{0.287217}$   \\
        \exercise{}    & \cite{lorenz2019}            & $\fnum{0.05}$ & $\fnum{0.212468}$   & $\ufnum{0.212972}$ & $\ufnum{0.363352}$   & $\fnum{0.432292}$  & $\ufnum{0.43724}$   & $\bfnum{0.331665}$  \\
        \exercise{}    & \cite{lorenz2019}            & $\fnum{0.1}$  & $\ufnum{0.218162}$  & $\fnum{0.203916}$  & $\fnum{0.305152}$    & $\fnum{0.43326}$   & $\fnum{0.429548}$   & $\fnum{0.318007}$   \\
        \hline
        \exercise{}    & Ours                         & -             & \bfnum{0.252595}    & \fnum{0.103587}    & \fnum{0.340158}      & \fnum{0.428429}    & \bfnum{0.504409}    & \ufnum{0.325835}    \\
        \midrule
    \end{tabular}
    \caption{Segmentation IOU for each unsupervised keypoint baseline + CRF \cite{Krahenbuhl2012EfficientInferenceFully} on each dataset. Bold means best, underlined means second best. The IOU performance strongly depends on the chosen variance of the isotropic Gaussian that models the extent of the keypoints. As one can see, taking a good hyperparameter from one dataset to another is not possible. Our proposed method does not rely on postprocessing and therefore is consistent across datasets.}
    \label{tab:completeioukeypoints}
\end{table*}

\label{sec:app_unsupervised_keypoint_baselines}

We highlight details on how we compare the results of unsupervised keypoint estimation baselines \cite{jakab2018unsupervised,lorenz2019,zhang2018unsupervised} with our results on unsupervised segmentation baselines.
We trained all unsupervised keypoint baseline with the same number of parts ($25$) and the same input image size ($128 \times 128$) as we trained our model.

\paragraph{Calibration Step}

The calibration step is depicted in \figref{fig:calibration_step}.
Given a set of inferred segmentations with paired ground-truth segmentations,
the IoU for all possible assignments from inferred segmentations to ground-truth segmentations is calculated.
Finally, the assignment with best IoU chosen.

\paragraph{Postprocessing of Keypoint Baselines}
To transform keypoints into segmentations, we create an isotropic Gaussian decay around each keypoint and then use it as unary potentials for a conditional random field inference. We noticed that the segmentation quality strongly depends on the variance $\sigma$ that is used the create the Gaussian decay.

We therefore optimized this parameter over a range of $\sigma$ parameters to find good segmentation in terms of final IOU with the ground-truth on the validation set.
To calculate the IOU, we apply the calibration step as described above.
We do this for each baseline and on each dataset individually and always select the \emph{best parameter configuration} based on overall IOU performance.
We do not apply any postprocessing on our results.

For completeness, we report the full table of segmentation results in \tabref{tab:completeioukeypoints}.
We observe that choosing a good hyperparameter for the variance of the Gaussian decay does not generalize across methods and datasets.
For \deepfashion{}{}, the variance in terms of IOU performance is very high.
We argue that this is because of the strong appearance variation of the dataset.
As already described, our method explicitly factors out some variation of the appearance by limiting the mutual information.
Therefore, our method works much better under such circumstances.
The effect is not as strong on exercise due to simple part appearances such as blue shirt, black sports pants, etc.
Therefore, a good segmentation can be achieved using only the CRF and keypoints.

\subsection{Additional Qualitative Results}

\begin{figure}[tb]
    \centering
    \def\ww{0.95}
    \def\www{0.95}
    \begin{tabular}{lc}
        Input                        & \parbox[c]{\ww\linewidth}{\includegraphics[width=\www\linewidth]{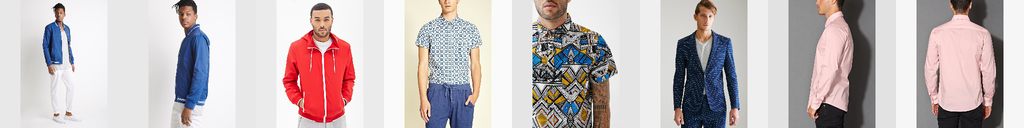}}      \\
        \hline
        ~                            & keypoints                                                                                                             \\
        \hline
        \cite{jakab2018unsupervised} & \parbox[c]{\ww\linewidth}{\includegraphics[width=\www\linewidth]{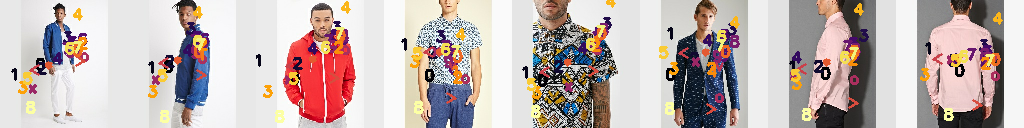}}   \\
        \cite{zhang2018unsupervised} & \parbox[c]{\ww\linewidth}{\includegraphics[width=\www\linewidth]{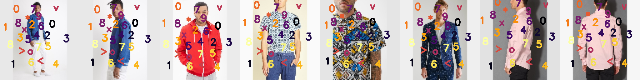}}   \\
        \cite{lorenz2019}            & \parbox[c]{\ww\linewidth}{\includegraphics[width=\www\linewidth]{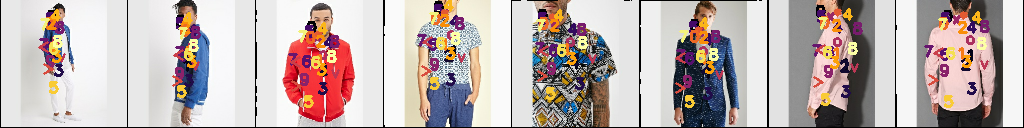}}  \\
        ~                            & keypoints + CRF postprocessing                                                                                        \\
        \hline
        \cite{jakab2018unsupervised} & \parbox[c]{\ww\linewidth}{\includegraphics[width=\www\linewidth]{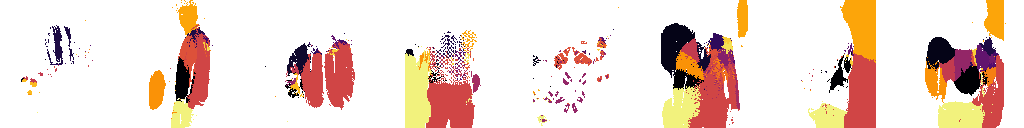}}  \\
        \cite{zhang2018unsupervised} & \parbox[c]{\ww\linewidth}{\includegraphics[width=\www\linewidth]{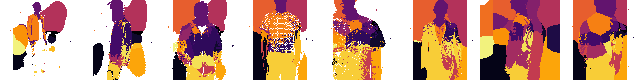}}  \\
        \cite{lorenz2019}            & \parbox[c]{\ww\linewidth}{\includegraphics[width=\www\linewidth]{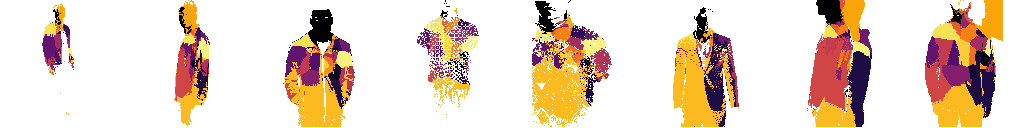}} \\
        ~                            & best assignment                                                                                                       \\
        \hline
        \cite{jakab2018unsupervised} & \parbox[c]{\ww\linewidth}{\includegraphics[width=\www\linewidth]{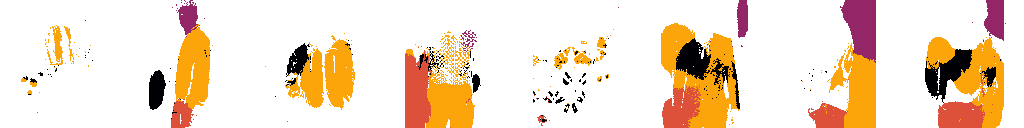}}        \\
        \cite{zhang2018unsupervised} & \parbox[c]{\ww\linewidth}{\includegraphics[width=\www\linewidth]{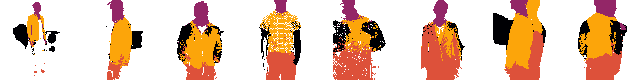}}        \\
        \cite{lorenz2019}            & \parbox[c]{\ww\linewidth}{\includegraphics[width=\www\linewidth]{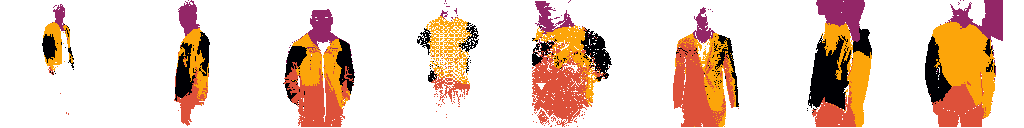}}       \\
        \hline
        Ours                         & \parbox[c]{\ww\linewidth}{\includegraphics[width=\www\linewidth]{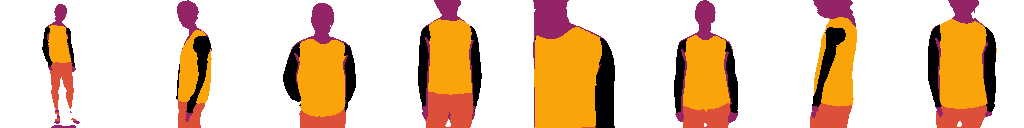}}       \\
        GT                           & \parbox[c]{\ww\linewidth}{\includegraphics[width=\www\linewidth]{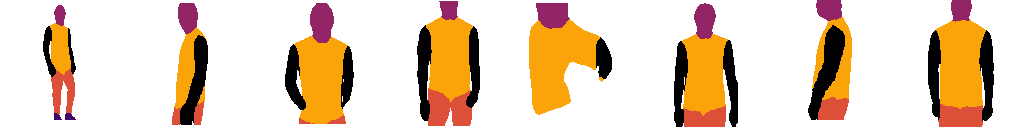}}
    \end{tabular}
    \caption{Additional results for comparison against keypoint-based baselines on DeepFashion.}
    \label{fig:additionalqualcompkeypointsdf}
\end{figure}

\begin{figure}[tb]
    \centering
    \def\ww{0.95}
    \def\www{0.95}
    \begin{tabular}{lc}
        Input                        & \parbox[c]{\ww\linewidth}{\includegraphics[width=\www\linewidth]{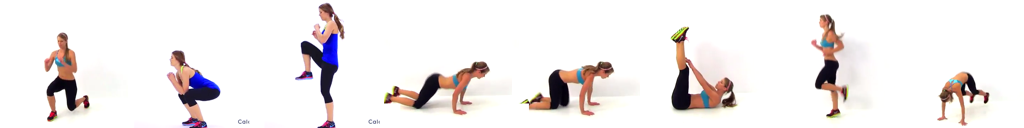}}         \\
        \hline
        ~                            & keypoints                                                                                                             \\
        \hline
        \cite{jakab2018unsupervised} & \parbox[c]{\ww\linewidth}{\includegraphics[width=\www\linewidth]{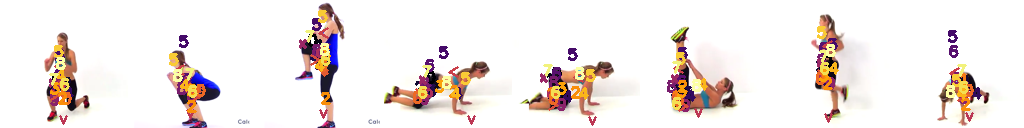}}   \\
        \cite{zhang2018unsupervised} & \parbox[c]{\ww\linewidth}{\includegraphics[width=\www\linewidth]{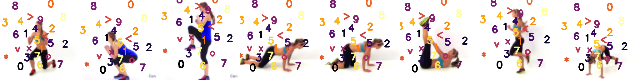}}   \\
        \cite{lorenz2019}            & \parbox[c]{\ww\linewidth}{\includegraphics[width=\www\linewidth]{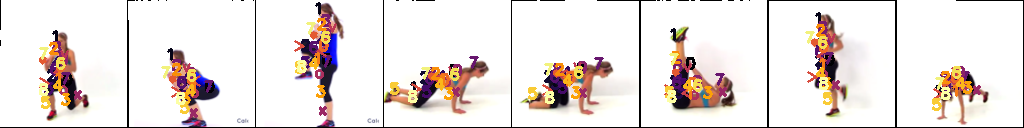}}  \\
        ~                            & keypoints + CRF postprocessing                                                                                        \\
        \hline
        \cite{jakab2018unsupervised} & \parbox[c]{\ww\linewidth}{\includegraphics[width=\www\linewidth]{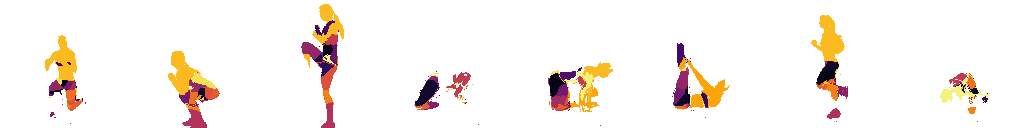}}  \\
        \cite{zhang2018unsupervised} & \parbox[c]{\ww\linewidth}{\includegraphics[width=\www\linewidth]{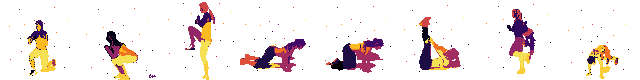}}  \\
        \cite{lorenz2019}            & \parbox[c]{\ww\linewidth}{\includegraphics[width=\www\linewidth]{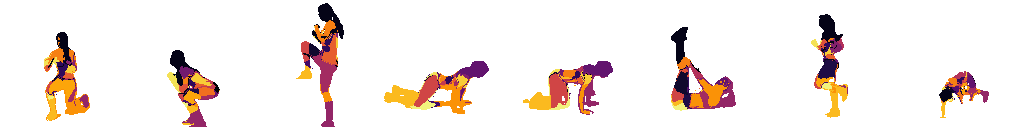}} \\
        ~                            & best assignment                                                                                                       \\
        \hline
        \cite{jakab2018unsupervised} & \parbox[c]{\ww\linewidth}{\includegraphics[width=\www\linewidth]{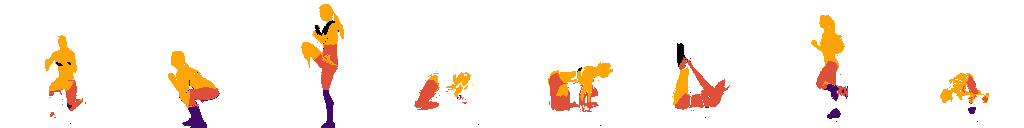}}        \\
        \cite{zhang2018unsupervised} & \parbox[c]{\ww\linewidth}{\includegraphics[width=\www\linewidth]{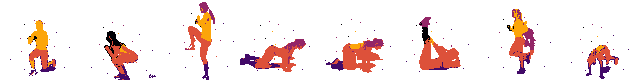}}        \\
        \cite{lorenz2019}            & \parbox[c]{\ww\linewidth}{\includegraphics[width=\www\linewidth]{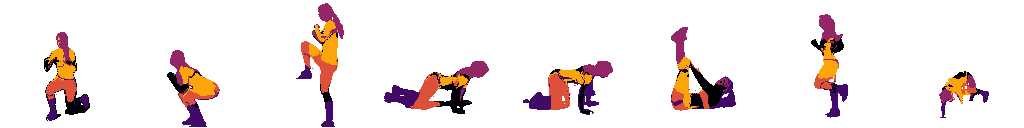}}       \\
        \hline
        Ours                         & \parbox[c]{\ww\linewidth}{\includegraphics[width=\www\linewidth]{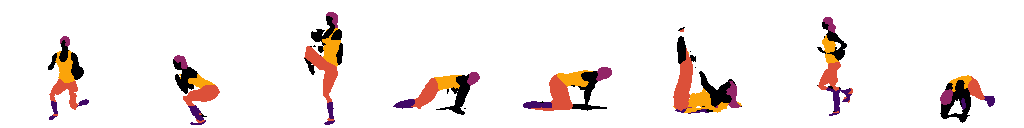}}          \\
        GT                           & \parbox[c]{\ww\linewidth}{\includegraphics[width=\www\linewidth]{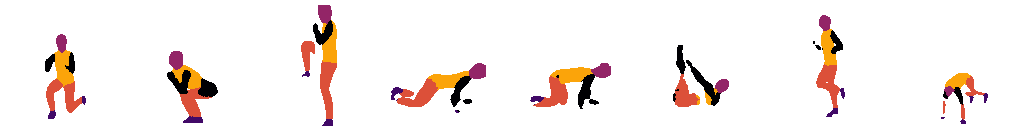}}
    \end{tabular}
    \caption{Additional results on \exercise{}.}
    \label{fig:additionalqualcompkeypointsex}
\end{figure}

\begin{figure}[tb]
    \centering
    \def\ww{0.95}
    \def\www{0.95}
    \begin{tabular}{lc}
        Input
         & \parbox[c]{\ww\linewidth}{\includegraphics[width=\www\linewidth]{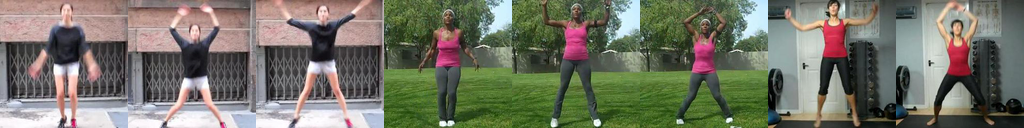}}
        \\
        \hline
        ~
         & keypoints
        \\
        \hline
        \cite{zhang2018unsupervised}
         & \parbox[c]{\ww\linewidth}{\includegraphics[width=\www\linewidth]{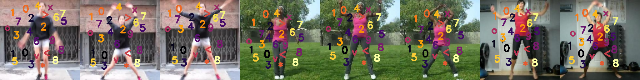}}
        \\
        \cite{jakab2018unsupervised}
         & \parbox[c]{\ww\linewidth}{\includegraphics[width=\www\linewidth]{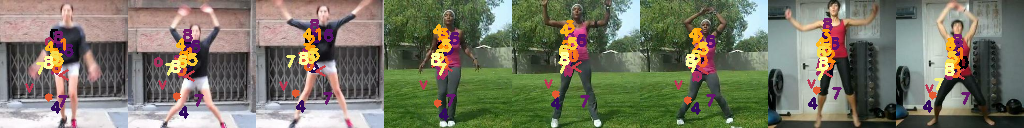}}
        \\
        \cite{lorenz2019}
         & \parbox[c]{\ww\linewidth}{\includegraphics[width=\www\linewidth]{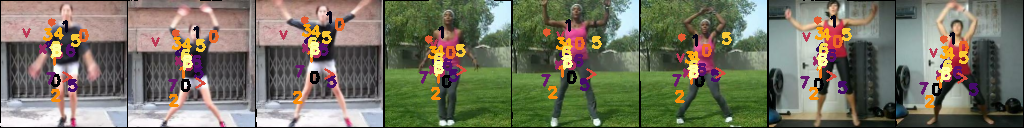}}
        \\
        ~
         & keypoints + CRF postprocessing
        \\
        \hline
        \cite{zhang2018unsupervised}
         & \parbox[c]{\ww\linewidth}{\includegraphics[width=\www\linewidth]{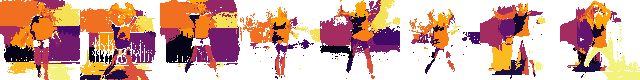}}
        \\
        \cite{jakab2018unsupervised}
         & \parbox[c]{\ww\linewidth}{\includegraphics[width=\www\linewidth]{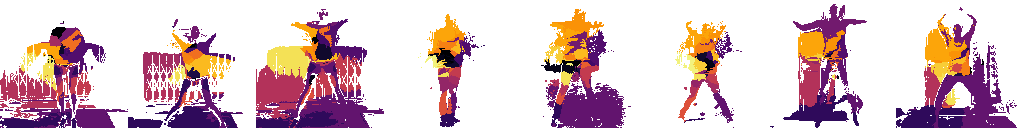}}
        \\
        \cite{lorenz2019}
         & \parbox[c]{\ww\linewidth}{\includegraphics[width=\www\linewidth]{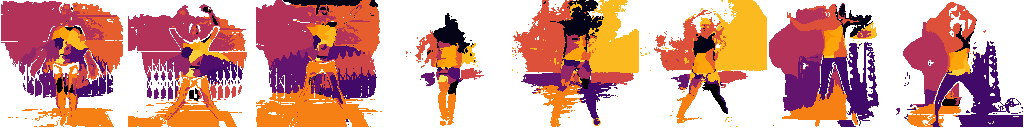}}
        \\
        ~
         & best assignment
        \\
        \hline
        \cite{zhang2018unsupervised}
         & \parbox[c]{\ww\linewidth}{\includegraphics[width=\www\linewidth]{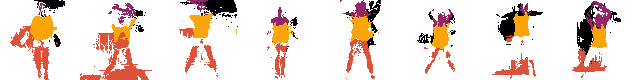}}
        \\
        \cite{jakab2018unsupervised}
         & \parbox[c]{\ww\linewidth}{\includegraphics[width=\www\linewidth]{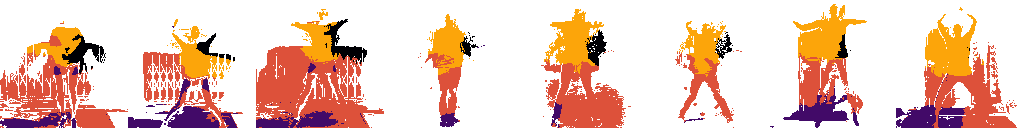}}
        \\
        \cite{lorenz2019}
         & \parbox[c]{\ww\linewidth}{\includegraphics[width=\www\linewidth]{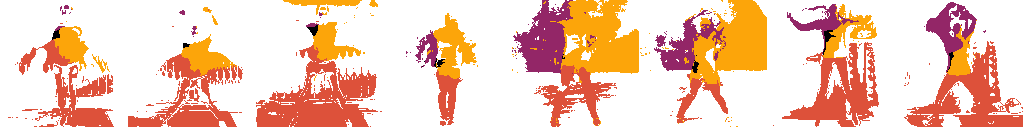}}
        \\
        \hline
        Ours
         & \parbox[c]{\ww\linewidth}{\includegraphics[width=\www\linewidth]{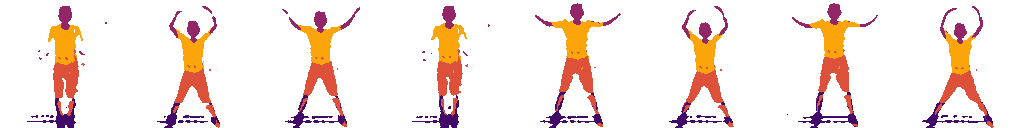}}
        \\
        GT
         & \parbox[c]{\ww\linewidth}{\includegraphics[width=\www\linewidth]{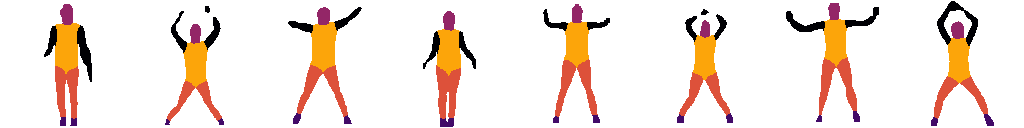}}
    \end{tabular}
    \caption{Additional results on \pennaction{}.}
    \label{fig:additionalqualcompkeypointspenn}
\end{figure}

We show additional qualitative results for comparison against keypoint-based baselines.
Results for \deepfashion{}{} are shown in \figref{fig:additionalqualcompkeypointsdf}, for \exercise{}{} in \figref{fig:additionalqualcompkeypointsex} and for \pennaction{} in \figref{fig:additionalqualcompkeypointspenn}.
It can clearly be seen that our method discovers parts which are consistent across instances and poses.
Furthermore, discovered keypoints are hard to assign to semantic parts.

\subsection{Derivation for Gaussian Markov Random Field}

For completeness, we derive \eqref{eq:klgmrf}.
For some parameter $\vec{y}$, a prior $p(\vec{y})$ is chosen as an improper GMRF with zero mean.
\begin{align}
    p(\vec{y})
     & = \normal{\vec{0}}{\vec{Q}^{-1}},
    ~ \vec{Q}^{\frac{1}{2}} = \nabla
\end{align}
Then, the posterior $p\left( \vec{y} \mid \vec{x} \right)$ is inferred from a different random variable $\vec{x}$.
In a variational inference framework, the posterior is now assumed to be an isotropic Gaussian around an inferred mean $f(\vec{x})$.
\begin{align}
    q\left( \vec{y} \mid \vec{x} \right)
     & = \normal{ f(\vec{x}) }{ \vec{I} }
\end{align}
By regularizing the $\dkl$ between $q$ and $p$, the posterior is kept close to the prior.
With the chosen $q$ and $p$, we get:
\begin{align}
    p(\vec{y})
                  & = \normal{\vec{0}}{\vec{Q}^{-1}},
    ~ \vec{Q}^{\frac{1}{2}} = \nabla
    \\
    q\left( \vec{y} \mid \vec{x} \right)
                  & = \normal{ f(\vec{x}) }{ \vec{I} }
    \\
    \mathcal{N}_0 & = \normal{\vec{\mu_0}}{\Sigma_0},~
    \mathcal{N}_1 = \normal{\vec{\mu_1}}{\Sigma_1}     \\
    \kldiv{
        \mathcal{N}_0
    }{
        \mathcal{N}_1
    }
                  & =
    \frac{1}{2} \left[
        \trace ( \Sigma_1^{-1} \Sigma_0 )
        +   (\vec{\mu}_1 - \vec{\mu}_0)^{\TT}
        \Sigma^{-1}_1
        (\vec{\mu}_1 - \vec{\mu}_0)^{\TT}
        + \mathrm{const}
    \right]                                            \\
    \kldiv{
        q
    }{
        p
    }
                  & =  \frac{1}{2} \left[
    \underbrace{\trace{(\vec{Q} \vec{I})}}_{\text{const}}
    + \underbrace{
        f(\vec{x})^{\TT}\vec{Q}f(\vec{x})
    }_{
    \norm{\vec{Q}^{\frac{1}{2}} f(\vec{x})}^2
    } + \mathrm{const}
    \right]
    = \lambda \norm{ \nabla f(\vec{x}) }^2 + \mathrm{const}.
\end{align}
Observe that the KL boils down to the gradient of $f(\vec{x})$, which in practice has to be approximated by finite differences.
If $f(\vec{x})$ is a 2D tensor for example, it is simply the image gradient.

\subsection{Part-based Apperance Transfer}

\begin{figure}[tb]
    \centering
    \parbox[c]{1\linewidth}{\includegraphics[width=1\linewidth]{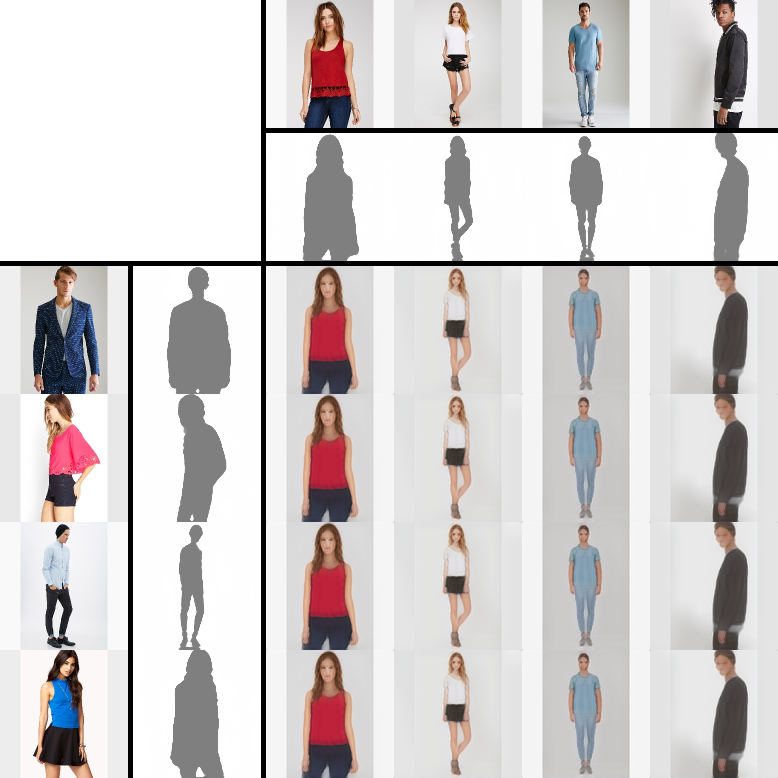}}
    \caption{
        \textbf{Part-based Appearance Transfer}.
        No Swaps applied.
    }
    \label{fig:transferhqa}
\end{figure}

\begin{figure}[tb]
    \centering
    \parbox[c]{1\linewidth}{\includegraphics[width=1\linewidth]{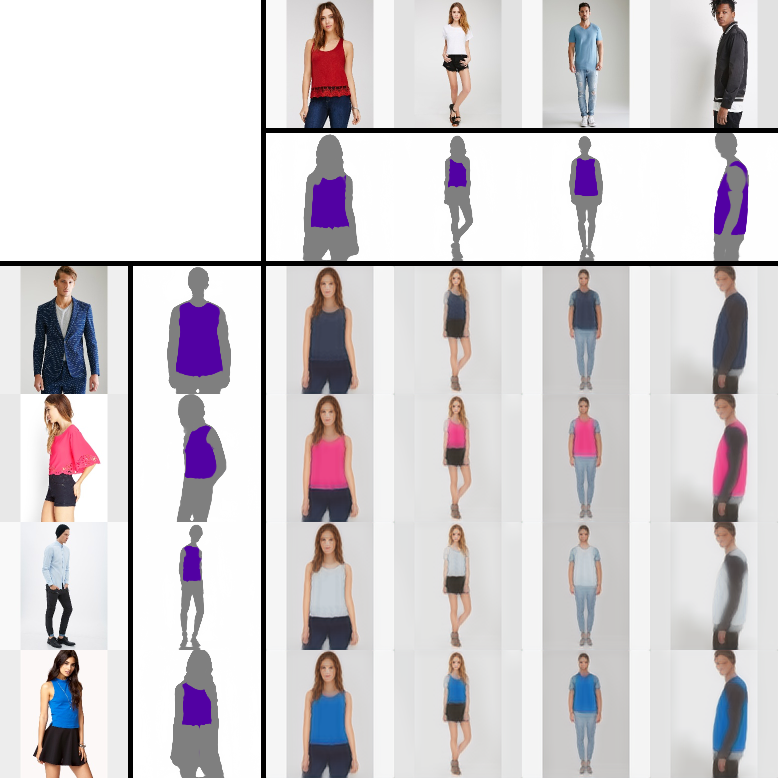}}
    \caption{
        \textbf{Part-based Appearance Transfer}.
        Swapping only chest appearance.
    }
    \label{fig:transferhqb}
\end{figure}

\begin{figure}[tb]
    \centering
    \parbox[c]{1\linewidth}{\includegraphics[width=1\linewidth]{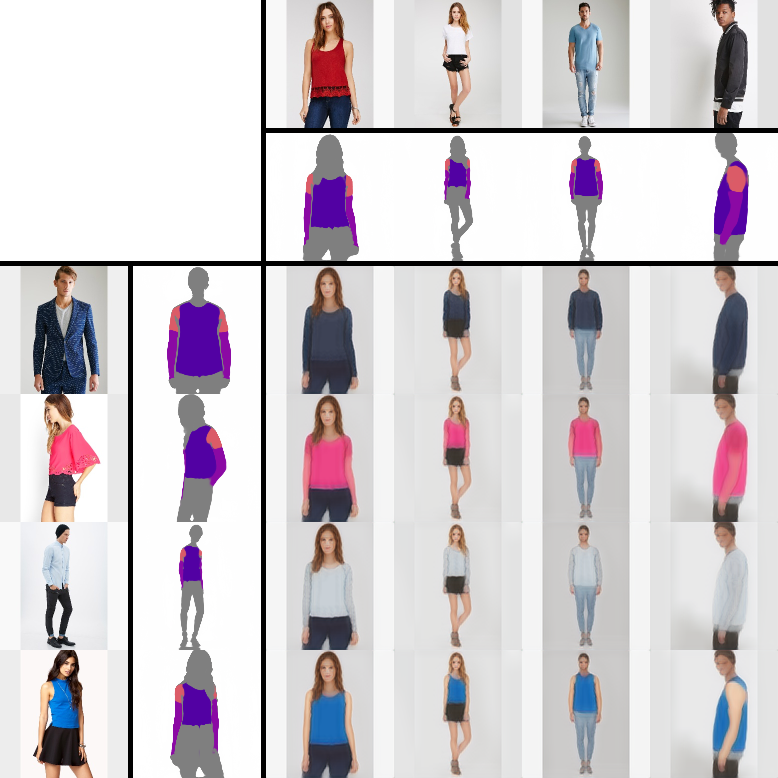}}
    \caption{
        \textbf{Part-based Appearance Transfer}.
        Swapping chest and arm appearance.
    }
    \label{fig:transferhqc}
\end{figure}

\begin{figure}[tb]
    \centering
    \parbox[c]{1\linewidth}{\includegraphics[width=1\linewidth]{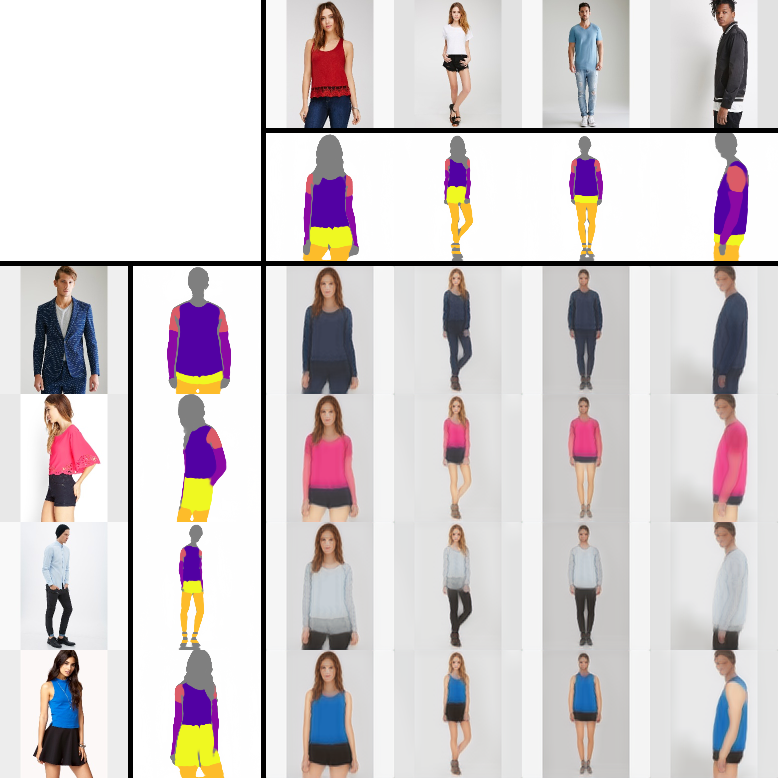}}
    \caption{
        \textbf{Part-based Appearance Transfer}.
        Swapping chest, arm, hip and leg appearance.
    }
    \label{fig:transferhqd}
\end{figure}

We show high resolution versions of \figref{fig:transfer} in \figref{fig:transferhqa}, \figref{fig:transferhqb}, \figref{fig:transferhqc}, \figref{fig:transferhqd}.


\begin{figure}[tb]
    \def\ww{0.1}
    \centering
    \null\hfill
    \begin{tabular}{c|c|c|c|ccc}
        ~
         & disentanglement
         & GMRF
         & $\loss{ \entropy{p} }$
         & \centered{\includegraphics[width = \ww\textwidth]{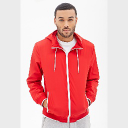}}
         & \centered{\includegraphics[width = \ww\textwidth]{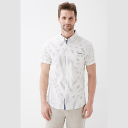}}
         & \centered{\includegraphics[width = \ww\textwidth]{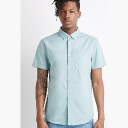}}
        \\ \midrule
        \centered{1.}
         & \centered{\xmark}
         & \centered{\xmark}
         & \centered{\xmark}
         & \centered{\skull}
         & \centered{\skull}
         & \centered{\skull}
        \\ \midrule
        \centered{2.}
         & \centered{variational                                                                                     \\+ \\ adversarial}
         & \centered{\xmark}
         & \centered{\xmark}
         & \centered{\includegraphics[width = \ww\textwidth]{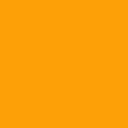}}
         & \centered{\includegraphics[width = \ww\textwidth]{figures/08_ablation/02_01.png}}
         & \centered{\includegraphics[width = \ww\textwidth]{figures/08_ablation/02_01.png}}
        \\ \midrule
        \centered{3.}
         & \centered{variational                                                                                     \\+ \\ adversarial}
         & \centered{\cmark}
         & \centered{\xmark}
         & \centered{\includegraphics[width = \ww\textwidth]{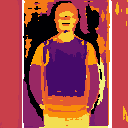}}
         & \centered{\includegraphics[width = \ww\textwidth]{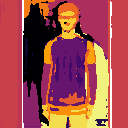}}
         & \centered{\includegraphics[width = \ww\textwidth]{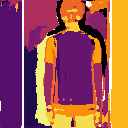}}
        \\ \midrule
        \centered{4.}
         & \centered{only                                                                                            \\ variational}
         & \centered{\cmark}
         & \centered{\cmark}
         & \centered{\includegraphics[width = \ww\textwidth]{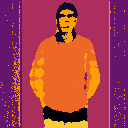}}
         & \centered{\includegraphics[width = \ww\textwidth]{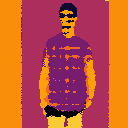}}
         & \centered{\includegraphics[width = \ww\textwidth]{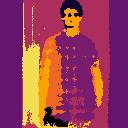}}
        \\ \midrule
        \centered{5.}
         & \centered{Full Model, variational                                                                         \\+ \\ adversarial}
         & \centered{\cmark}
         & \centered{\cmark}
         & \centered{\includegraphics[width = \ww\textwidth]{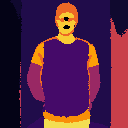}}
         & \centered{\includegraphics[width = \ww\textwidth]{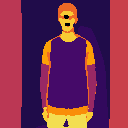}}
         & \centered{\includegraphics[width = \ww\textwidth]{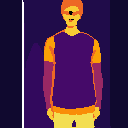}}
    \end{tabular}
    \hfill\null
    \caption{
        \textbf{Analyzing Disentanglement, GMRF prior and Entropy Regularization}.
        We highlight the importance of each introduced prior in a series of ablation studies.
    }
    \label{fig:ablationa}
\end{figure}

\subsection{Ablation Studies}

We conduct an ablation study in \figref{fig:ablationa}.
To start simple, we use a model without disentanglement, GMRF prior or entropy regularization.
We expect this model to fail, since there is no incentive to factorize the distribution into independent shape and appearance representations. The model diverges after roughly $\num{3000}$ steps.

We introduce disentanglement by adding variational and adversarial objectives.
This prevents the model from diverging, however it converges to a an undesired local minimum consisting of a single constant part.

We now constrain the solutions of $S$ by adding the GMRF prior, which clearly encourages localized parts to be discovered.

Adding the entropy regularization objective increases the smoothness of our solution.

Finally we ask the question if we really need variational and adversarial objectives or if either of them is sufficient.
We expect this model to take into account appearance cues such as average color, instead of semantic consistency across instances when factorizing the data distribution.
The experiment validates this hypothesis as the torso parts are not consistently labelled as the same part, which previously has been the case.

%% file: figures/svg/calibration.pdf_tex
\begingroup%
  \makeatletter%
  \providecommand\color[2][]{%
    \errmessage{(Inkscape) Color is used for the text in Inkscape, but the package 'color.sty' is not loaded}%
    \renewcommand\color[2][]{}%
  }%
  \providecommand\transparent[1]{%
    \errmessage{(Inkscape) Transparency is used (non-zero) for the text in Inkscape, but the package 'transparent.sty' is not loaded}%
    \renewcommand\transparent[1]{}%
  }%
  \providecommand\rotatebox[2]{#2}%
  \newcommand*\fsize{\dimexpr\f@size pt\relax}%
  \newcommand*\lineheight[1]{\fontsize{\fsize}{#1\fsize}\selectfont}%
  \ifx\svgwidth\undefined%
    \setlength{\unitlength}{1111.79573119bp}%
    \ifx\svgscale\undefined%
      \relax%
    \else%
      \setlength{\unitlength}{\unitlength * \real{\svgscale}}%
    \fi%
  \else%
    \setlength{\unitlength}{\svgwidth}%
  \fi%
  \global\let\svgwidth\undefined%
  \global\let\svgscale\undefined%
  \makeatother%
  \begin{picture}(1,0.4092497)%
    \lineheight{1}%
    \setlength\tabcolsep{0pt}%
    \put(0,0){\includegraphics[width=\unitlength,page=1]{calibration.pdf}}%
    \put(0.64671062,0.30314235){\color[rgb]{0,0,0}\makebox(0,0)[lt]{\lineheight{1.25}\smash{\begin{tabular}[t]{l}part assigment 1: IoU 0.3\end{tabular}}}}%
    \put(0.64268488,0.1771824){\color[rgb]{0,0,0}\makebox(0,0)[lt]{\lineheight{1.25}\smash{\begin{tabular}[t]{l}part assigment 2: IoU 0.7\end{tabular}}}}%
    \put(0.63912357,0.04583233){\color[rgb]{0,0,0}\makebox(0,0)[lt]{\lineheight{1.25}\smash{\begin{tabular}[t]{l}part assigment 3: IoU 0.6\end{tabular}}}}%
    \put(0,0){\includegraphics[width=\unitlength,page=2]{calibration.pdf}}%
    \put(0.76327545,0.1230196){\color[rgb]{0.66666667,0.53333333,0}\makebox(0,0)[lt]{\lineheight{1.25}\smash{\begin{tabular}[t]{l}Final assignment\\after calibration\end{tabular}}}}%
    \put(0.04490466,0.10855065){\color[rgb]{0,0,0}\makebox(0,0)[t]{\lineheight{1.25}\smash{\begin{tabular}[t]{c}inferred\\parts\end{tabular}}}}%
    \put(0.30752167,0.39593367){\color[rgb]{0,0,0}\makebox(0,0)[t]{\lineheight{1.25}\smash{\begin{tabular}[t]{c}GT\\parts\end{tabular}}}}%
  \end{picture}%
\endgroup%